\documentclass[10pt]{article}
\usepackage[accepted]{tmlr}

\usepackage{hyperref}
\usepackage{url}
\renewcommand{\cite}{\citep}
\usepackage{multirow}

\usepackage{lipsum}
\usepackage{amsfonts, amsmath, amssymb, amsthm}
\usepackage{mathrsfs, mathtools, bbm, bm}
\usepackage[capitalize, nameinlink]{cleveref}
\usepackage{graphicx}
\usepackage{epstopdf}
\usepackage{algorithmic}
\ifpdf
  \DeclareGraphicsExtensions{.eps,.pdf,.png,.jpg}
\else
  \DeclareGraphicsExtensions{.eps}
\fi

\newtheorem{remark}{Remark}[section]
\newtheorem{theorem}{Theorem}[section]
\newtheorem{definition}{Definition}[section]
\newtheorem{corollary}{Corollary}[section]

\crefname{hypothesis}{Hypothesis}{Hypotheses}

\newtheorem{example2}{Example}
\crefname{example2}{Example}{Example}
\crefformat{equation}{(#2#1#3)}
\newcommand{\crefrangeconjunction}{--}
\crefrangeformat{equation}{(#3#1#4)\crefrangeconjunction(#5#2#6)}
\crefname{app}{Appendix}{Appendices}
\Crefname{app}{Appendix}{Appendices}

\newcommand{\surr}[1]{\tilde{#1}}
\newcommand{\surrbest}[1]{\hat{#1}}
\newcommand{\inputdim}{{D}}
\newcommand{\outputdim}{{Q}}
\newcommand{\N}{\mathbb{N}}
\newcommand{\R}{\mathbb{R}}

\newcommand{\surrclass}{\mathcal{F}}
\renewcommand{\P}{\mathbb{P}}
\newcommand{\ndata}{K}
\newcommand{\nparams}{P}
\newcommand{\generator}{\bar{g}}
\renewcommand{\vec}[1]{\bm{#1}}
\newcommand{\ip}[2]{\langle #1, #2 \rangle}
\renewcommand{\d}{\text{d}}
\newcommand{\loss}{\mathcal{L}}
\DeclareMathOperator*{\Var}{Var}

\DeclareMathOperator*{\argmin}{arg \, min}

\usepackage{verbatim}
\usepackage{float}
\usepackage{upgreek}

\title{SUPN: Shallow Universal Polynomial Networks for Efficient Function Approximation}

\author{\name Zachary Morrow\thanks{These authors contributed equally to this work.}            \email zbmorro@sandia.gov \\
    \addr Department of Scientific Machine Learning \\
    Sandia National Laboratories
    \AND
    \name Michael Penwarden$^*$ \email mspenwa@sandia.gov \\
    \addr Center for Monitoring Systems and Technology \\
    Sandia National Laboratories
    \AND
    \name Brian Chen \email brichen@sandia.gov \\
    \addr Department of Optimization and Uncertainty Quantification \\
    Sandia National Laboratories
    \AND
    \name Aurya Javeed \email asjavee@sandia.gov \\
    \addr Department of Optimization and Uncertainty Quantification \\
    Sandia National Laboratories
    \AND
    \name Akil Narayan \email akil@sci.utah.edu \\
    \addr Department of Mathematics \\
    Scientific Computing and Imaging Institute \\
    University of Utah
    \AND
    \name John D. Jakeman \email jdjakem@sandia.gov \\
    \addr Department of AI Credibility \\
    Sandia National Laboratories
}

\def\month{10}
\def\year{2026}
\def\openreview{\url{https://openreview.net/forum?id=srHGkCrSTM}}

\begin{document}

    \maketitle

    \begin{abstract}
        Deep neural networks (DNNs) and Kolmogorov--Arnold networks (KANs) provide flexible function approximators, but their accuracy can require substantial network depth, width, and trainable parameter counts. Beyond making the resulting networks less transparent, large parameter spaces can increase training cost and produce solutions that are sensitive to optimization and parameter initialization, leading to variability in generalization error. For these reasons, we propose Shallow Universal Polynomial Networks (SUPNs), which retain a single hidden layer while increasing the expressivity of each unit through a learnable multivariate polynomial inside each activation, which we refer to as a polynomial pre-activation. This construction trades architectural depth and width for greater expressivity within each hidden unit, reducing parameter counts and training cost while preserving a shallow computational graph. We establish two complementary universal approximation results: SUPNs are universal by increasing network width at fixed polynomial complexity, as in classical shallow-network approximation, and by increasing polynomial complexity at fixed finite width. The latter provides a distinct route to increasing approximation capacity, and our timing results show that increasing polynomial complexity produces comparatively little growth in training cost. We further prove that finite SUPNs can approach the best approximation attainable in a prescribed polynomial space to arbitrary relative tolerance, derive explicit quasi-optimal parameters, and establish corresponding convergence rates. We empirically evaluate SUPNs against DNNs, KANs, and polynomial projection on smooth, nonsmooth, and discontinuous function-approximation problems in one, two, and ten dimensions using more than 16,000 trained models. Across the target functions studied, SUPNs frequently achieve lower approximation error and variability than DNNs and KANs at comparable parameter counts while retaining lower training cost, supporting their use as efficient function approximators when the target structure is not known \textit{a priori}.
    \end{abstract}

    \section{Introduction}
\label{sec:introduction}

Many scientific applications involve computationally expensive input--output maps, denoted $f$, resulting from physical systems, such as state-space evolution~\cite{PEHERSTORFER2016196} or the relationship between a model's parameters and certain quantities of interest~\cite{morrow2020method}. Accordingly, much work has been devoted toward constructing less costly approximations equipped with error estimates or convergence guarantees. One approach is focused on accelerating a numerical solver, such as a finite-element method, by simplifying a model's governing equations~\cite{jakeman2025evaluation} or constructing a reduced-order model~\cite{benner2017model} that diminishes the number of unknowns. Though this approach is focused on the physical system and not directly on $f$, such improvements will nonetheless reduce the expense of constructing $f$. Another approach is to approximate $f$ directly. In practice, this requires sampling its inputs and outputs, and an approximation constructed this way is called a {\em data-driven surrogate model}. Throughout this work, we use \emph{function approximation} to describe the mathematical problem and \emph{surrogate modeling} when emphasizing its use as a computational replacement for the original model. The efficiency of this approach depends on the dimensionality and nonlinearity of $f$. Many methods in this category are concerned primarily with the high-dimensional case because a system may have a very large number of parameters mapped to some quantity of interest. In this paper, we present a highly expressive method for approximating nonlinear and low-regularity functions that is consistently competitive with, and often outperforms, DNNs, KANs, and polynomial approximation across a broad set of benchmark problems. For transparency and to avoid the known issue of weak baselines in machine learning research~\cite{mcgreivy2024weak}, we compare our method to deep neural networks (DNNs), Kolmogorov--Arnold networks (KANs), and polynomial projection on an extensive set of numerical experiments with documented hyperparameter choices.

Several different varieties of data-driven surrogate models exist, and we will purposefully save the technical formulation for a later section. Although an exhaustive list is beyond the scope of this paper, we offer a brief description. A data-driven surrogate of an input--output map can be broadly categorized as linear or nonlinear, depending on how the parameters appear within the surrogate's ansatz. Linear methods express the resultant surrogate as a linear combination of chosen basis functions. This category includes polynomial chaos expansions \cite{Ghanem1991, doi:10.1137/S1064827501387826}, sparse grids \cite{doi:10.1137/060663660,Barthelmann2000}, operator inference \cite{PEHERSTORFER2016196,doi:10.1137/130932715}, spectral methods \cite{etde_21021742,doi:10.1137/1.9781611970425}, and radial basis functions \cite{Buhmann_2003}. In contrast, the parameters of nonlinear methods are embedded within the surrogate's constituent functions, which are generally nonlinear. The parameters of a nonlinear surrogate are determined with an iterative optimization procedure. Functional tensor trains \cite{gorodetsky2019continuous} and DNNs are examples of this category.

Although linear surrogates are well-established methods with a rich body of theory, DNNs have attracted significant contemporary interest due to their flexibility and expressivity, resulting in a proliferation of architectures \cite{cybenko1989approximation,6795963,ronneberger2015u,liu2025kan,vaswani2017attention} and bespoke adaptations tailored to specific problems, such as AlphaFold \cite{Jumper2021}. Nonetheless, the search for a sufficiently accurate DNN is often hampered by vanishing or exploding gradients, an unwieldy number of trainable parameters, and difficulty approximating high-frequency target functions \cite{zhang2025why}. Theoretical results establish that feedforward networks, also known as multi-layer perceptrons (MLPs), can approximate continuous functions arbitrarily well \cite{pinkus1999approximation}. An alternative to the MLP, KANs are based on the Kolmogorov--Arnold representation theorem \cite{Arnold2009,kolmogorov1957representations,braun2009constructive}. In contrast to MLPs, KANs have learnable parameters within their activation functions, which are parametrized as, e.g., splines \cite{splines} or Chebyshev polynomials \cite[p.~211]{atkinson1989introduction}. In fact, KANs with a linear spline basis are equivalent to MLPs with a ReLU activation function \cite{actor2025leveraging}. Neural networks (NNs) have also been extended to learn discretizations of operators between function spaces. Kernel-based methods such as the Fourier neural operator (FNO) \cite{li2021fourier} and kernel neural operator (KNO) \cite{lowery2026kernel} seek to learn a sequence of kernel integral operators. Deep operator networks (DeepONets) \cite{lu2021learning} are an alternative approach consisting of two NNs, one learning a set of basis functions and the other learning a set of coefficients.

Our motivation is to trade network depth and width for greater expressivity within each hidden unit. A conventional shallow MLP applies a nonlinear activation to an affine function of the inputs, whereas we instead consider applying the activation to a learnable multivariate polynomial. This provides a second mechanism for increasing approximation capacity beyond adding depth or width while retaining a single hidden layer. As shown in Section~3, this construction retains universal approximation, while the stronger quantitative results relate its approximation error directly to classical polynomial best approximation. Thus, the polynomial pre-activation provides additional per-neuron expressivity while retaining a shallow computational graph and a tractable connection to established approximation theory.

Motivated by this tradeoff, we introduce a single-layer $\tanh$-activated network with an initial lift consisting of Chebyshev polynomials, which we call a {\em shallow universal polynomial network} (SUPN). This function-approximation method leverages the power of both polynomial approximation and NNs (via a nonlinear activation function) to produce an approximation that is both parsimonious and robust with respect to network parameter initialization. Indeed, the principal theoretical contributions of this paper (\cref{ssec:poly-supn}) go beyond standard universal approximation theorems and establish that SUPNs converge at the same rate as a related polynomial approximation. Our numerical experiments empirically indicate that SUPNs are competitive against DNNs and KANs on smooth target functions, and they are competitive against polynomial projection on nonsmooth targets, although our theory does not rigorously establish {\em superior} performance over projection in the nonsmooth case. Moreover, we observed that, without fail, SUPNs perform at least as well as---and often much better than---DNNs or KANs when the target function has tensor-product structure.

\begin{figure}[htbp!]
    \centering
    \includegraphics[width=1\textwidth]{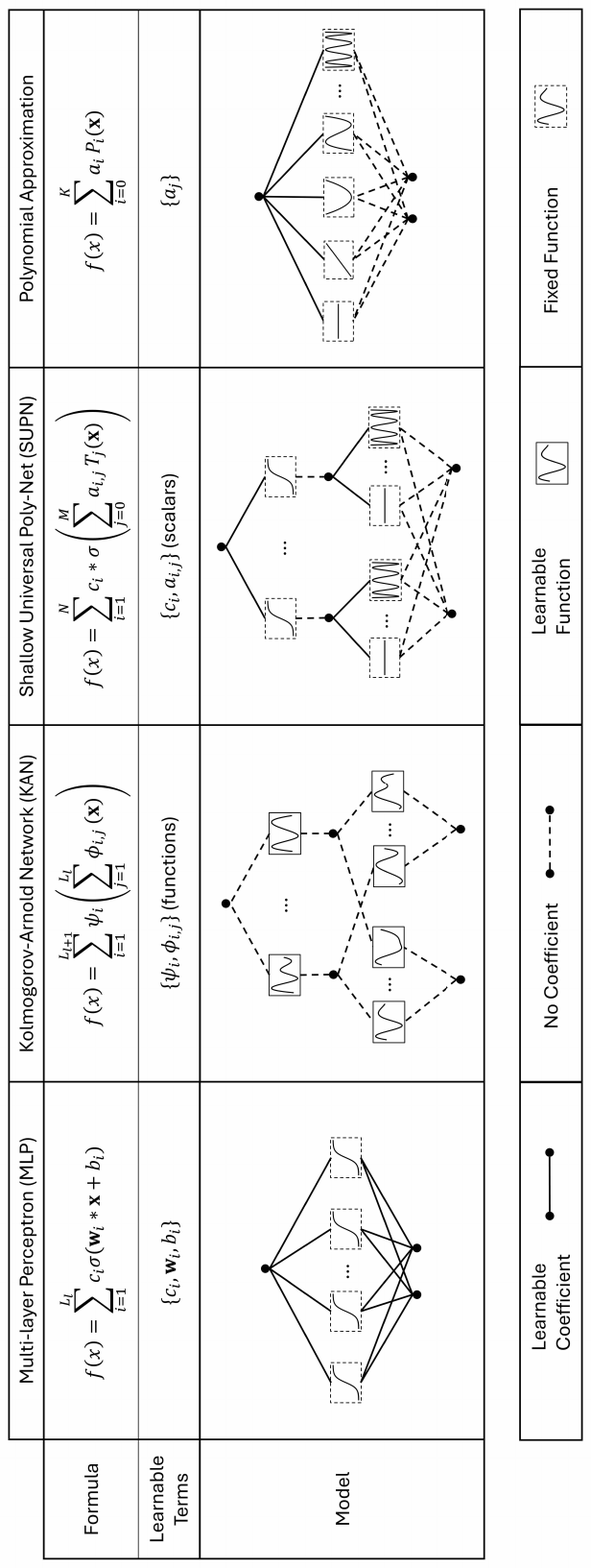}
    \caption{Illustration of function approximators considered in this work. Note that MLPs and KANs are shown in their single layer form for comparison, but are most often used in a deep (many-layer) configuration compared to SUPNs, which are always a single layer.}
    \label{fig:architectures}
\end{figure}

\Cref{fig:architectures} shows a visual comparison of SUPNs with other methods. SUPNs differ from MLPs through the multivariate polynomial lift preceding each activation, and from KANs through both their shallow structure and the use of a multivariate polynomial within each neuron rather than compositions of learned univariate edge functions. In particular, previous discussions of Chebyshev KANs \cite{ss2024chebyshev,shukla2024comprehensive} adopted a deep-learning approach that composed tanh-activated Chebyshev layers together. In addition, \citet{xu2025chebyshev} proposed a Chebyshev feature map where the learnable parameters are not basis coefficients but instead generalized polynomial ``degrees''; those networks still use a deep MLP after the Chebyshev lift.

The term \textit{polynomial networks} has historically referred to methods that utilize polynomial activation functions. Despite the similarity in naming conventions, our method has little in common with these prior models, such as polynomial neural networks (PNNs)~\cite{DAS1995231, Kak1993, doi:10.1073/pnas.79.8.2554}, deep PNNs~\cite{kileel2019expressive, chrysos2021deep, kubjas2024geometry}, and sum-product networks~\cite{poon2011sum}. What unites these methods is that they are reformulations or extensions of the so-called Group Method of Data Handling \cite{4308320}, which builds hierarchical polynomial regressions and is equivalent to an MLP with polynomial activation functions. However, it is well-established that MLPs with polynomial activation functions are not universal approximators \cite{LESHNO1993861,sonoda2017neural, pinkus1999approximation}. In contrast, our proposed method, based on linear combinations of $\tanh$-activated polynomials, {\em is} a universal approximator.

To motivate the derivation of SUPNs, enable their analysis, and demonstrate their approximation capabilities, the rest of the paper is organized as follows. \Cref{sec:background} provides a general framework for function approximation and error estimation, as well as basic formulations of polynomial projection, MLPs, and KANs. \Cref{sec:supn} presents the mathematical formulation of SUPNs, establishes two basic universal approximation theorems, and proves SUPN convergence rates based on polynomial error bounds. \Cref{sec:optimizer} discusses the second-order optimizer utilized for training. \Cref{sec:numerics} compares SUPNs, polynomial projection, DNNs, and KANs in one, two, and ten dimensions on standard test functions (e.g., Runge and Rastrigin) and specific examples from KAN papers. Existing literature currently lacks a systematic comparison of machine learning methods against polynomial approximations. \Cref{sec:conclusion} summarizes our findings and discusses possible extensions of this work. The appendices summarize the notation used throughout this paper, present supplementary numerical results, and provide supporting technical details of proofs.

    \section{Function approximation} \label{sec:background}

Consider a model $f$ mapping $\inputdim$ bounded inputs to $\outputdim$ outputs. We can denote this model as $f: \R^\inputdim \supset \Omega \to \R^\outputdim$ where $\Omega \subset \R^D$ is compact.\footnote{Without loss of generality, we may take $\Omega = [-1, 1]^\inputdim$ since a compact interval $[a, b]$ can be mapped to $[-1, 1]$ with a straightforward affine transformation. We also often take $Q=1$ for convenience, since the multi-output case constructs a scalar-valued result for each output  dimension.} The first step of function approximation is to define a space $\surrclass$ of candidate functions, parameterized by an ansatz $\generator = \generator(\cdot \, ; \vec{\theta})$ and $\vec{\theta} \in \R^\nparams$ such that $\mathcal{F}_{\vec{\theta}} = \{ \generator(\cdot \, ; \vec{\theta}) : \vec{\theta} \in \R^P \}$. At the highest level, the goal of function approximation is to find a suitable approximation $\surrbest{f} \in \surrclass_{\vec{\theta}}$. With $p \geq 1$ and $\mu$ a positive measure on $\Omega$, the best approximation of $f$ by $\surrclass_{\vec{\theta}}$ in the $\mu$-weighted $L^p(\Omega)$ norm is
\begin{align}
\surrbest{f} &\coloneqq \argmin_{g \in \mathcal{F}_{\vec{\theta}}} \| f-g \|_{L^p_\mu} \, .
\label{eq:best_surrogate_F}
\end{align}
A minimum exists (rather than an infimum) because $\mathcal{F}_{\vec{\theta}}$ is finite-dimensional and the norm is a continuous function bounded below. An alternative form of~\cref{eq:best_surrogate_F}, which will become useful later, is
\begin{equation}
    \surrbest{f} = \generator(\cdot \, ; \vec{\surrbest{\theta}}), \qquad \vec{\surrbest{\vec{\theta}}} = \argmin_{\vec{\theta} \in \R^P} \| f - \generator (\cdot \, ; \vec{\theta}) \|_{L^p_\mu} \, .
    \label{eq:best_surrogate}
\end{equation}
The choice of norm and $\generator$ in~\cref{eq:best_surrogate} are crucial to the accuracy of $\surrbest{f}$. Common choices of $\generator$ are degree-$n$ polynomials (denoted $\P_n$), sines and cosines with maximum frequency $n$, or neural networks of a given width and depth. Well-established theory exists for these ansatzes, and we will review some of this theory later in this section. 

In practice, surrogates are constructed instead from a finite amount of data because the explicit functional form of $f$ is rarely available. In the supervised learning context, one collects $K \in \N$ input--output pairs $\{ (\vec{x}_k, f(\vec{x}_k)) \}_{k \in [\ndata]}$, where $[\ndata] \coloneqq \{ 1, \dots, \ndata \}$. Data-driven surrogates approximate the continuous $L^p$ norms with a corresponding discrete $\ell^p$ norm. The discrete minimization problem becomes
\begin{equation}
    \begin{split}
        \surr{f} &= \generator(\cdot \, ; \vec{\surr{\theta}}) , \qquad \qquad \hspace{19pt} \vec{\surr{\theta}} = \argmin_{\vec{\theta} \in \R^\nparams} \| \vec{\generator}(\vec{\theta}) - \vec{f} \|_{\ell^p_{\vec{w}}(\R^\ndata)}, \\
        \vec{\generator}(\vec{\theta}) &= \left(\generator(\vec{x}_k; \vec{\theta})\right)_{k \in [\ndata]}, \qquad \vec{f} = \left(f(\vec{x}_k)\right)_{k \in [\ndata]},
    \end{split}
\label{eq:finitedim_surrogate}
\end{equation}
which is a direct adaptation of the best-approximation problem~\cref{eq:best_surrogate}. The weights $\vec{w} \in \R^K$ are chosen so that the discrete
$\ell^p_{\vec{w}}$ norm approximates the corresponding continuous $L^p$
norm. For $p<\infty$, this amounts to approximating
\[
\|v\|_{L^p(\Omega)}^p
=
\int_\Omega |v(x)|^p \,\d x
\]
with a quadrature rule:
\[
\|v\|_{L^p(\Omega)}^p
\approx
\sum_{k\in[K]} w_k |v(x_k)|^p
=
\| \vec{v} \|_{\ell^p_{\vec{w}}(\R^K)}^p.
\]
Thus, the sampling nodes and weights determine an empirical approximation of
the continuous norm used in the best-approximation problem:
$$
\| \vec{v} \|_{\ell^p_{\vec{w}}(\R^K)} = \begin{cases}
    \left( \sum_{k \in [K]} w_k |v_k|^p \right)^{1/p}, &\qquad \mbox{if }p < \infty \\
    \max_{k \in [K]} |v_k|, &\qquad \mbox{if }p = \infty
\end{cases} \ .
$$

The final step is to minimize~\cref{eq:finitedim_surrogate}. Popular optimization methods include grid search and gradient-based optimizers. Grid searches discretize the feasible set for $\vec{\theta}$ and take the discrete minimum. However, grid searches often do not provide optimality guarantees, and $P$ is typically quite large, so the number of samples required to cover the sample space (e.g., with low discrepancy) is practically prohibitive. Alternatively, when using gradient-based optimizers, the $\ell^p$ norm should be differentiable. As a result, practical implementation uses~\cref{eq:finitedim_surrogate} with the $\ell^2$ norm, yielding the familiar loss-minimization problem with mean-squared error (MSE):
\begin{equation}
    \vec{\surr{\theta}} = \argmin_{\vec{\theta} \in \R^\nparams} \loss(\vec{\theta}), \qquad \loss(\vec{\theta}) = \sum_{k \in [\ndata]} w_k \left( \bar{g}(\vec{x}_k; \vec{\theta}) - f(\vec{x}_k) \right)^2 \, .
    \label{eq:loss}
\end{equation}
In the special case where $\generator$ depends linearly on $\vec{\theta}$, the minimizer $\vec{\surr{\theta}}$ is the solution to a linear system, which is significantly cheaper to solve than the nonlinear case.

\subsection{Ansatzes}

We now present three prominent choices of the ansatz $\generator$. These ansatzes have associated theory that will be crucial in the theoretical analysis of \cref{sec:supn}, and they will be used for comparison in \cref{sec:numerics}.

\subsubsection{Polynomials} \label{ssec:projection}

Consider how to approximate a function $f: \Omega \to \R$ with polynomials in some norm. For this paper, there are two important cases to consider: the $L^\infty$ norm and the $L^2$ norm. The first case is addressed in the following theorems, which will be needed in the proofs of \cref{sec:supn}.

\begin{theorem}[\citet{stone1937applications, weierstrass1885analytische}]
    Let $f: \Omega \to \R$ be continuous and $\Omega \subset \R^D$ be compact. For any $\epsilon > 0$, there exists a polynomial $q : \Omega \to \R$ such that $\| f - q \|_{L^\infty(\Omega)} < \epsilon$.
    \label{thm:stone-weierstrass}
\end{theorem}

\begin{theorem}[\citet{jackson1930theory}] \label{thm:jackson}
    Let $f \in C^k([-1, 1], \R)$ with $f^{(k)}$ $\alpha$-H\"older continuous, i.e. there exists $\gamma > 0$ and $\alpha \in (0, 1]$ such that $|f^{(k)}(x) - f^{(k)}(y)| \leq \gamma |x-y|^{\alpha}$. There exists a constant $\beta > 0$ such that, for every $n \geq 1$, there is a polynomial $q_n \in \P_n$ for which
    \begin{equation*}
        \| f - q_n \|_{L^\infty} \leq \beta \, n^{-(k+\alpha)} \ .
    \end{equation*}
\end{theorem} 

Now consider a polynomial approximation of $f \in L^2(\Omega)$ with degree at most $\nparams$ in the unweighted $L^2$ norm. For simplicity, we address $D=1$. This corresponds to~\cref{eq:best_surrogate} with the $L^2$ norm and
\begin{equation}
\bar{g}_P(x ; \vec{\theta}) = \sum_{p \in [\nparams-1]_0} \theta_p \, L_{p}(x),
\label{eq:legendre-series}
\end{equation}
where $[\nparams-1]_0 \coloneqq \{ 0, \dots, \nparams-1\}$ and $L_p$ is the Legendre polynomial of degree $p$. The Legendre polynomials satisfy
\begin{align*}
\ip{L_n}{L_m}_{L^2(\Omega)} &= \delta_{nm} \, \frac{2}{2n+1} \, .
\end{align*}
where
\begin{align*}
      \ip{f}{g}_{L^2(\Omega)} &\coloneqq \int_{\Omega} f(x) g(x) \, \d x \, , & \| f\|_{L^2(\Omega)}^2 &\coloneqq \ip{f}{f}_{L^2(\Omega)} \, .
\end{align*}
In this case, the coefficients of the best approximation are
\begin{equation*}
\surrbest{\theta}_p = \frac{\ip{L_p}{f}_{L^2(\Omega)}}{\ip{L_p}{L_p}_{L^2(\Omega)}} = \ip{L_p}{f}_{L^2(\Omega)}\, \frac{2p+1}{2} \, ,
\end{equation*}
and we have
\begin{equation*}
\lim_{P \to \infty} \| f - \bar{g}_P \|_{L^2} = 0 \, .
\end{equation*}
The data-driven setting approximates $\ip{L_p}{f}_{L^2(\Omega)}$ with a quadrature rule
\begin{equation}
    \surrbest{\theta}_p \approx \surr{\theta}_p = \frac{2p+1}{2} \ \sum_{k \in [\ndata]} w_k L_p(x_k) f(x_k) \, .
    \label{eq:datadriven_projection_coefs}
\end{equation}

\paragraph{Projection in higher dimensions} \label{ssec:projection_multidim}
It will later become necessary to consider $D \geq 2$. In that case, the notion of ``degree'' must be generalized to allow for differing amounts of cross-interaction terms among the input space. This is accomplished by replacing the sum over $p \in [\nparams-1]_0$ with a sum over $\vec{p} \in \Lambda \subset \N_0^d$, where $\Lambda$ is a lower set and the multi-dimensional Legendre polynomials are products of the univariate ones:
\begin{equation*}
L_{\vec{p}}(\vec{x}) = \prod_{d=1}^D L_{p_d}(x_d) \, .
\end{equation*}
In this case, the approximation space becomes
$$
\P_{\Lambda} = \text{span} \{ L_{\vec{p}} (\vec{x}): \vec{p} \in \Lambda \} \ .
$$

\begin{definition}[Lower set~\cite{chkifa2018polynomial}]
    A set $\Lambda \subset \N_0^D$ is called lower (or downward-closed) if and only if, for each $\vec{i} \in \Lambda$,  $\{ \vec{j}: \vec{j} \leq \vec{i} \} \subset \Lambda$. Here, $\vec{j} \leq \vec{i}$ if and only if $j_d \leq i_d$ for all $d \in [D]$.
    \label{def:lower}
\end{definition}

\begin{figure}
    \centering
    \includegraphics[width=0.7\textwidth]{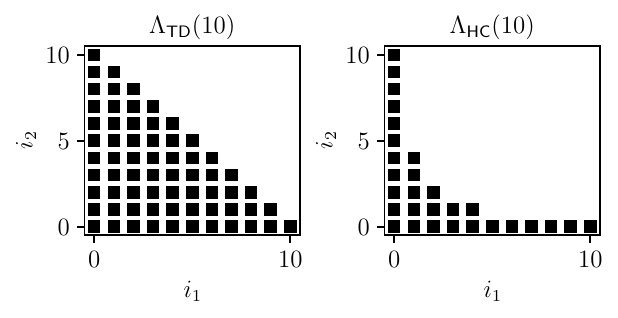}
    \caption{Two examples of lower sets, the total-degree space (left) and hyperbolic cross-section (right).}
    \label{fig:lower-set}
\end{figure}

\begin{example2}
    Two common examples of lower sets are the hyperbolic-cross and total-degree spaces, shown in \Cref{fig:lower-set} and given by, respectively,
    \begin{align}
        \Lambda_{\text{HC}}(M) &= \left\{ \vec{i} \in \N_0^D : \prod_{d=1}^D (i_d + 1) \leq M+1 \right\}, \label{eq:lambda-hc} \\
        \Lambda_{\text{TD}}(M) &= \left\{ \vec{i} \in \N_0^D: \sum_{d=1}^D i_d \leq M \right\}. \label{eq:lambda-td}
    \end{align}
\end{example2}

\subsubsection{Multi-layer perceptrons} \label{ssec:mlp}
A multi-layer perceptron (MLP) is given by
\begin{equation}
\begin{cases}
    ~~\surr{f}(\vec{x}) &= \vec{W}_L \vec{y}_L(\vec{x}) \\
    ~~\vec{y}_{k+1}(\vec{x}) &= \sigma(\vec{W}_{k} \vec{y}_{k}(\vec{x}) + \vec{b}_{k}), \qquad k \in [L-1] \\
    ~~\vec{y}_1(\vec{x}) &= \sigma(\vec{W}_0 \, \vec{x} + \vec{b}_0)
\end{cases} \, .
\label{eq:mlp}
\end{equation}
Here, $L$ is the network depth, $N = \dim(\vec{b}_k)$ is the uniform network width, and $\sigma(x)$ is a non-polynomial activation function applied component-wise. The network parameters $\vec{\theta}$ are the concatenation of the entries of each $\vec{W}_k$ and $\vec{b}_k$. By counting the size of each $\vec{W}_k$ and $\vec{b}_k$, the total number of parameters is $P = N(D+2) + (L-1)(N^2+N)$.

We now state two universal approximation theorems for MLPs that will be foundational for the proofs in \cref{sec:supn}.

\begin{theorem}[Shallow, wide MLPs~\cite{pinkus1999approximation}]
Let both $f: \Omega \to \R^Q$ and $\sigma: \R \to \R$ be continuous. Let the network depth be fixed at $L=1$. Then $\sigma$ is non-polynomial if and only if, for any $\epsilon > 0$, there exists $N \in \N$, $\vec{W}_0 \in \R^{N \times D}$, $\vec{b}_0 \in \R^N$, and $\vec{W}_1 \in \R^{Q \times N}$ such that $\tilde{f}$ in~\cref{eq:mlp} satisfies
\begin{equation*}
\sup_{\vec{x} \in \Omega} \| f(\vec{x}) - \tilde{f}(\vec{x}) \|_{\ell^\infty(\R^Q)} < \epsilon
\end{equation*}
where $\sigma(\cdot)$ is applied componentwise.
\label{thm:arbitrary-width}
\end{theorem}

\begin{theorem}[Deep, narrow MLPs~\cite{kidger2020universal}]
Let $f : \Omega \to \R^Q$ be continuous, and let $\sigma : \R \to \R$ be non-affine. Let the network width $N$ be fixed such that $N \geq D+Q+2$. Then there exists $L \in \N$, weights $\{ \vec{W}_k \}_{k \in [L]_0}$, and biases $\{ \vec{b}_k \}_{k \in [L-1]_0}$ such that $\surr{f}(\vec{x})$ in~\cref{eq:mlp} satisfies
\begin{equation*}
\sup_{\vec{x} \in \Omega} \| f(\vec{x}) - \surr{f}(\vec{x}) \|_{\ell^\infty(\R^Q)} < \epsilon \, .
\end{equation*}
\label{thm:arbitrary-depth}
\end{theorem}

\subsubsection{Kolmogorov--Arnold networks} \label{ssec:kan}

A deep KAN \cite[Eq. 2.10--2.12]{liu2025kan} has the form
\begin{equation}
    \surr{f}(\vec{x})
    =
    \psi_{L} \circ \psi_{L-1} \circ \cdots \circ \psi_{0}(\vec{x}),
    \label{eq:kan}
\end{equation}
where the KAN layer
$\psi_\ell : \R^{N_\ell} \to \R^{N_{\ell+1}}$
is defined as
\begin{equation}
    (\psi_\ell(\vec{z}))_k
    =
    \left(
        \vec{W}_{\ell}\,\text{SiLU}(\vec{z})
    \right)_k
    +
    \sum_{j \in [N_\ell]}
    \phi_{\ell,k,j}^{(p)}(z_j),
    \label{eq:kan_layer}
\end{equation}
where $\ell \in [L]_0$, $k \in [N_{\ell+1}]$, and
$\vec{W}_\ell \in \R^{N_{\ell+1} \times N_\ell}$. In the original formulation of KANs, the functions $\phi$ are univariate B-splines of the form
\begin{equation*}
    \phi^{(p)}(x)
    =
    \sum_{n \in [G]} \alpha_n B_{n,p}(x),
\end{equation*}
where $G$ is the size of the grid partition
$\{x_n\}_{n \in [G]}$ and $p$ is the spline order.
The splines themselves are defined recursively via
\begin{equation*}
    B_{n,0}
    \coloneqq
    \begin{cases}
        1, & x_n \leq x < x_{n+1}, \\
        0, & \text{otherwise},
    \end{cases}
\end{equation*}
and, for $p > 0$,
\begin{equation*}
    B_{n,p}(x)
    \coloneqq
    \frac{x-x_n}{x_{n+p}-x_n}B_{n,p-1}(x)
    +
    \frac{x_{n+p+1}-x}{x_{n+p+1}-x_{n+1}}
    B_{n+1,p-1}(x).
\end{equation*}
The network parameters $\vec{\theta}$ are the union of all
$\vec{W}_\ell$ and all B-spline coefficients.

\subsection{Empirical accuracy estimation} \label{ssec:accuracy-estimation}

To evaluate the effectiveness a function approximation, we identify some useful metrics such as \textit{best approximation error}, \textit{sampling error}, and \textit{training robustness}. We will utilize these metrics in \cref{sec:numerics}. For clarity of presentation, we take $D=1$ here.

\subsubsection{Best approximation error}
This error is determined purely by the choice of $\vec{\theta} \in \R^\nparams$. The error under consideration is
\begin{equation}
  \epsilon(\vec{\surrbest{\theta}}) \coloneqq \inf_{g \in \surrclass} \| f-g\|_{L^p_\mu} = \| f - \surrbest{f} \|_{L^p_\mu} \, ,
  \label{eq:best-approx-error}
\end{equation}
where $\surrbest{f}$ is defined in~\cref{eq:best_surrogate}. Crucially, $\epsilon(\vec{\surrbest{\theta}})$ is free from any sampling or training error.

\subsubsection{Sampling error}
This error is due to the replacement of the continuous $L^2$ norm in~\cref{eq:best_surrogate} with an empirical norm. Using this norm, we seek to minimize the empirical training loss~\cref{eq:loss}. The corresponding generalization error is given by
\begin{equation*}
    \epsilon(\surr{\vec{\theta}}) = \| f - \tilde{f} \|_{L^2_\mu}
\end{equation*}
where $\surr{f}$ is given by~\cref{eq:finitedim_surrogate}. The meaningful indicator about the effect of sampling would be a comparison between $\epsilon(\surrbest{\vec{\theta}})$ and $\epsilon(\surr{\vec{\theta}})$. Fixing $K$, we examine three potential choices of grids on $[-1, 1]$, where $k \in [\ndata]$:
    \begin{itemize}
        \item {\em Uniform sampling}: 
        \begin{align*}
            x_k &\stackrel{\text{iid}}{\sim} \mathcal{U}([-1, 1]), & w_k &= 2/K
        \end{align*}
        \item {\em Equidistant sampling}:
        \begin{align*}
            x_k &= -1 + \frac{2(k-1)}{K-1}, & w_k &= 2/K
        \end{align*}
        \item {\em Gauss--Legendre quadrature}~\cite[p. 276]{atkinson1989introduction}:
        \begin{align*}
            \{ x_k \}_{k \in [\ndata]} &= L_{K}^{-1}(\{ 0 \}), & w_k &= \frac{-2}{(K+1)L_{K+1}(x_k) L_K'(x_k)}
        \end{align*}
    \end{itemize}

\subsubsection{Training error and robustness}  \label{sssec:training-error}
When numerically optimizing the parameters of a nonlinear function approximator, typically we can only discover various local minima of~\cref{eq:loss} instead of a global minimum. However, different local minima of the training loss can result in dramatically different generalization error. Premature stopping of the optimizer further compounds this issue (see \cref{sec:optimizer}). Much attention has been given to the proper initialization of network weights for desirable generalization error, e.g. through explicit probability distributions \cite{cyr2020robust, fokina2020growing, lee2024improved, lu2020dying, glorot2010understanding, he2015delving} or \textit{meta-learned} initializations \cite{finn2017model,nichol2018first,liu2022novel,penwarden2023metalearning}. We examine training robustness by randomly initializing the network many times and reporting the variance of the test error:
\begin{equation*}
    \Var_{\vec{\theta}_\text{init}\sim \mu} \left[ \epsilon(\vec{\surr{\theta}}) \right]
\end{equation*}
where $\mu$ is the Kaiming uniform distribution~\cite{he2015delving}.

    \section{Shallow universal polynomial networks (SUPNs)} \label{sec:supn}

For one input dimension, we define a {\em shallow universal polynomial network} as any function of the form
\begin{align}\label{eq:ansatz}
  f_{N,M}(x) \coloneqq \sum_{n \in [N]} c_n \tanh\left( \sum_{m \in [M-1]_0} a_{n,m} \, T_{m}(x) \right), \qquad x \in [-1, 1],
\end{align}
where $c_n$, and $a_{n,m}$ are trainable parameters and $T_m(x)$ is the degree-$m$ Chebyshev polynomial.  Intuitively, a SUPN replaces the first $L-1$ layers of a deep MLP with a Chebyshev polynomial, yielding $P=N(M+1)$ . This reduces the number of trainable network parameters while retaining an explicit multivariate polynomial inside each hidden activation. For our purposes, $T_m$ may be computed either trigonometrically or via the three-term recurrence relation~\cite[p.~211]{atkinson1989introduction}. In principle, any polynomial basis on $\Omega$ could be used in~\cref{eq:ansatz}, but we observed that the Chebyshev basis performs well in practice. We hypothesize that the special approximation properties of Chebyshev polynomials may be responsible for this performance. For instance, among all degree-$n$ polynomials with unit $L^\infty$ norm, $T_n$ has largest leading coefficient, and each local extremum is $\pm 1$, which provides stability when learning $a_{n,m}$. Again, SUPNs differ from Chebyshev KANs through the absence of repeated compositions.

In higher dimensions, we define a multi-dimensional SUPN as
\begin{equation}
    f_{N,\Lambda}(\vec{x}) \coloneqq \sum_{n \in [N]} c_n \tanh\left( \sum_{\vec{m} \in \Lambda} a_{n,\vec{m}} \, T_{\vec{m}}(\vec{x}) \right)
    \label{eq:ansatz_multidim}
\end{equation}
where $\Lambda \subset \N_0^D$ is a lower set and $T_{\vec{m}}(\vec{x}) = \prod_{d=1}^D T_{m_d}(x_d)$, as in \cref{ssec:projection_multidim}.
In this multi-dimensional setting, the number of trainable parameters for a SUPN is $P=N(|\Lambda|+1)$; for convenience, we may use $M$ and $|\Lambda|$ interchangeably when context is clear. Multi-dimensional SUPNs can model cross-interactions between inputs within a single layer through the multivariate polynomial basis. As with other multivariate polynomial approximations, however, the cardinality of $\Lambda$ can grow
rapidly with input dimension, so SUPNs do not remove the curse of dimensionality for arbitrary multivariate functions. Their scalability depends on whether the target admits an efficient representation within a sparse or structured multi-index set. Hyperbolic-cross and other downward-closed spaces can substantially reduce the number of retained cross-interaction terms relative to a full tensor-product space, particularly for functions with anisotropic or low effective dimensional structure \cite{cohen2015approximation}. In this paper, we use isotropic choices of $\Lambda$; adaptive anisotropic index sets are left for future work.

\begin{remark}[Multiple outputs]
    For clarity of presentation, this paper considers only the scalar-output case. However, SUPNs can straightforwardly handle multiple outputs by inserting an additional index on the inner and outer indices corresponding to the output dimension. This is mathematically equivalent to separate scalar SUPNs for each output, but the implementation will be more efficient due to vectorization.
\end{remark}

\subsection{Universal approximation of SUPNs} \label{ssec:uat}

We first establish universal approximation of the SUPN function class \eqref{eq:ansatz} through two short results based on established approximation theorems for neural networks and polynomials. These results verify that the shallow architecture with polynomial pre-activations retains universal approximation and provide the foundation for the stronger finite-complexity approximation results in Section \ref{ssec:poly-supn}.

\begin{theorem}[fixed $M$, variable $N$]
\label{thm:fixed_M}
Let $f : \Omega \to \R^Q$ be continuous with $\Omega \subset \R^D$ compact. Fix $M \geq D+1$. For any $\epsilon > 0$, there exists $N \in \N$ such that $\| f - f_{N, M} \|_{L^\infty} < \epsilon$.
\end{theorem}

\begin{proof}
    Let $\epsilon > 0$ be arbitrary. By \Cref{thm:arbitrary-width}, there exists $N \in \N$, $\vec{W}_0 \in \R^{N \times D}$, $\vec{b}_0 \in \R^N$, and $\vec{W}_1 \in \R^{Q \times N}$ such that $\| f - \surr{f}_N \|_{L^\infty} < \epsilon$, where
    \begin{equation*}
    \surr{f}_N(\vec{x}) = \vec{W}_1 \tanh( \vec{W}_0 \vec{x} + \vec{b}) \, ,
    \end{equation*}
    with $\tanh(\cdot)$ applied componentwise. Each entry of $\vec{W}_0 \vec{x} + \vec{b}$ is a linear polynomial and can be exactly represented by the inner sum of~\eqref{eq:ansatz_multidim} with $\{ 1, x_1, \dots, x_D \}$. This basis corresponds to $\Lambda = \{ \vec{0}, \vec{e}_1, \dots, \vec{e}_D \}$, giving $M = |\Lambda| = D+1$. $\hfill \square$
\end{proof}

\begin{theorem}[fixed $N$, variable $M$]
\label{thm:fixed_N}
Let $f : \Omega \to \R^Q$ be continuous, with $\Omega \subset \R^D$ compact. Fix $N \geq D+Q+2$. For any $\epsilon > 0$, there exists $M \in \N$ such that $\| f - f_{N, M} \|_{L^\infty} < \epsilon$.
\end{theorem}

\begin{proof}
    Fix $N \geq D+Q+2$, and let $\epsilon > 0$. By \Cref{thm:arbitrary-depth}, there exists $L \in \N$, $\{ \vec{W}_k \}_{k\in [L]_0}$, and $\{ \vec{b}_k \}_{k \in [L-1]_0}$ such that 
    \begin{equation*}
    \sup_{\vec{x} \in \Omega} \| f(\vec{x}) - \surr{f}_L(\vec{x}) \|_{\ell^\infty(\R^Q)} < \epsilon/2,
    \end{equation*}
    where $\surr{f}_L$ is given by~\cref{eq:mlp}. Let $\surr{f}_{L-1}(\vec{x})$ denote the output of layer, i.e. $\surr{f}_L(\vec{x}) = \vec{W}_L \tanh\left( \surr{f}_{L-1}(\vec{x}) \right)$. By \Cref{thm:stone-weierstrass} and the uniform continuity of $\tanh$ and $\surr{f}_{L-1}$, there exists a multi-index set $\Lambda$ and a vector-valued polynomial $p \in \P_{\Lambda}(\Omega; \R^Q)$ such that, for all $\vec{x} \in \Omega$, 
    \begin{equation*}
    \| f_{N, M}(\vec{x}) - \tilde{f}_L(\vec{x}) \|_{\ell^\infty(\R^Q)} < \epsilon / 2\, ,
    \end{equation*}
    where $f_{N, M}(\vec{x}) = \vec{W}_L \tanh(p(\vec{x}))$ and $M = |\Lambda|$. The triangle inequality completes the proof.
    $\hfill \square$
\end{proof}

\begin{remark}
    In principle, any non-polynomial activation function $\sigma(x)$ would suffice in \Cref{thm:fixed_M} and \Cref{thm:fixed_N} for universal approximation. However, there are practical and theoretical reasons for using $\tanh$. In practice, we observed numerical instabilities when $\sigma(x)$ is not bounded, such as the ReLU activation. Furthermore, the relationship between SUPN and polynomial approximation accuracy in \Cref{prop:poly-approx} relies on a Taylor expansion whose remainder is bounded by the derivatives of $\tanh$.
\end{remark}
\addtocounter{proof2}{1}

\subsection{Polynomial approximation bounds}
\label{ssec:poly-supn}

Universal approximation establishes density of the SUPN function class as
model complexity increases, but it does not quantify the approximation error attainable at a prescribed complexity. Quantitative approximation results also exist for other neural architectures; for example, deep ReLU networks admit Sobolev-class approximation bounds~\cite{yarotsky2017error}, while residual KANs admit Besov approximation rates~\cite{kratsios2026approximation}. Here,
we derive a different finite-complexity guarantee by relating SUPNs directly to classical polynomial best approximation. \Cref{prop:poly-approx} and \Cref{cor:poly-approx} show that, for a prescribed polynomial space $\P_\Lambda$ with best-approximation error $\epsilon_\Lambda(f)$, a finite SUPN can approach this best-approximation error to arbitrary relative tolerance with a comparable number of parameters. This relationship is specific to the SUPN construction and is not implied by standard universal approximation or
existing neural-network approximation-rate results. \Cref{thm:constructive-L2,thm:constructive-Linf} further make this relationship constructive by deriving explicit quasi-optimal SUPN parameters in the $L^2_\mu$ and $L^\infty$ norms. Finally, part~(b) of
\Cref{thm:constructive-Linf} provides a best-approximation convergence rate determined by the smoothness of $f$. These results concern best approximation within the SUPN function class at finite model complexity and should be distinguished from finite-sample generalization bounds, which additionally depend on the sampling process and learning procedure.

We state the following results in the weighted Sobolev space
$W^{k,p}_\mu(\Omega)$, consisting of functions whose weak derivatives through
order $k$ belong to $L^p_\mu(\Omega)$:
\begin{equation*}
W^{k,p}_\mu(\Omega) = \{ f : \| f\|_{W^{k,p}_\mu(\Omega)} < \infty \}, \qquad
\|f\|_{W^{k,p}_\mu}
\coloneqq
\max_{\substack{|\vec{\beta}|\leq k\\
\vec{\beta}\in\N_0^D}}
\left\|
\frac{\d^{|\vec{\beta}|}f}
{\d x_1^{\beta_1}\cdots\d x_D^{\beta_D}}
\right\|_{L^p_\mu(\Omega)}.
\end{equation*}
\Cref{prop:poly-approx} relates SUPN approximation to the best polynomial
approximation of $f\in W^{k,p}_\mu(\Omega)$ in the $W^{k,p}_\mu$ norm and
shows that a near-best polynomial can itself be approximated arbitrarily
closely by a SUPN. Its proof motivates the use of
$\sigma(x)=\tanh(x)$ as our activation function.
\Cref{cor:poly-approx} then gives the corresponding finite-SUPN
best-approximation guarantee. \Cref{thm:constructive-L2} derives explicit
quasi-optimal parameters in the $L^2_\mu$ norm, while
\Cref{thm:constructive-Linf} specializes the construction to the
$L^\infty$ norm and establishes the corresponding convergence rate.

\begin{theorem}[Polynomial approximation proximity] \label{prop:poly-approx}
  Let $f \in W^{k,p}_\mu(\Omega;\R)$, with $\mu$ a probability measure, $\Omega \subset \R^D$ compact, $k \in \N_0$, and $1 \leq p \leq \infty$. For $\Lambda \subset \N_0^D$ and with $\P_\Lambda \coloneqq \mathrm{span}_{\vec{m}\in \Lambda} T_{\vec{m}}$, define
    \begin{align*}
      \epsilon_{\Lambda}(f) \coloneqq \inf_{q \in \P_\Lambda} \|f - q\|_{W^{k,p}_\mu}.
    \end{align*} 
    For any $\delta > 0$, there is a polynomial $\tilde{q} \in \P_{\Lambda}$ and a SUPN of the form \eqref{eq:ansatz_multidim} with $P = |\Lambda|+1$ parameters such that  
    \begin{alignat}{2}
        &\left\| f - \tilde{q} \right\|_{W^{k,p}_\mu} &&< \left\{\begin{array}{rl} 
            (1 + \delta) \epsilon_{\Lambda}(f), & \textrm{ if }~\epsilon_{\Lambda}(f) > 0 \\
            \delta, & \textrm{ if }~\epsilon_{\Lambda}(f) = 0
            \end{array}\right. \qquad \text{and}  \label{eq:qtilde} \\[0.5em]
        &\left\|f_{1,\Lambda} - \tilde{q} \right\|_{W^{k,\infty}} &&< \left\{\begin{array}{rl} 
            \delta \, \epsilon_{\Lambda}(f), & \textrm{ if }~\epsilon_{\Lambda}(f) > 0 \\
            \delta, & \textrm{ if }~\epsilon_{\Lambda}(f) = 0
        \end{array}\right. \quad .\label{eq:supn-q}
    \end{alignat}
\end{theorem}

\begin{proof} \label{proof:poly-approx}
  We deal explicitly with $k = 0$, $\epsilon_{\Lambda}(f) > 0$. An analogous proof holds for $\epsilon_{\Lambda}(f) = 0$. We give the proof for $k > 0$ in \cref{sec:sm-sobolev-theory}. By definition of $\epsilon_\Lambda(f)$, there is some polynomial $\tilde{q} \in \P_\Lambda$ satisfying \cref{eq:qtilde}.  Let $\tilde{q} = \sum_{\vec{m} \in \Lambda} \alpha_{\vec{m}} T_{\vec{m}}$, and define $R \coloneqq \sum_{\vec{m} \in \Lambda} |\alpha_{\vec{m}}|$. If $R = 0$, set $c_1 = 0$ in \eqref{eq:ansatz_multidim} with $N = 1$ to achieve $f_{N,\Lambda} = q = 0$, achieving \eqref{eq:supn-q}. Otherwise, consider $R > 0$. Note that,
  \begin{align*}
    \sup_{\vec{x} \in \Omega} |\tilde{q}(\vec{x})| \leq \sup_{\vec{x} \in \Omega} \sum_{\vec{m} \in \Lambda} |\alpha_{\vec{m}}| |T_{\vec{m}}(\vec{x})| \leq R \, .
  \end{align*}
  We will construct a SUPN with $(N,M) = (1,|\Lambda|)$ that well-approximates $\tilde{q}$. Let $\sigma(y) = \tanh(y)$. Around $y = 0$, $\sigma(y) \approx y$, and using $|\sigma''(y)| < 2 |y|$, Taylor's Theorem with the integral form of the remainder gives, for any $r > 0$,
  \begin{align*}
    \left| \sigma(y) - y \right| &< r^3, & |y| &< r \, .
  \end{align*}
  Now consider the $(N,M) = (1,|\Lambda|)$ SUPN in \eqref{eq:ansatz_multidim}, with the parameter assignment
  \begin{align*}
    a_{1,\vec{m}} &= \frac{\alpha_{\vec{m}}}{S}, & c_1 &= S, & S&\coloneqq \sqrt{\frac{R^3}{\delta\epsilon_\Lambda(f)}} \, .
  \end{align*}
  Then, for any $\vec{x} \in \Omega$, 
  \begin{align*}
    \left|f_{N,\Lambda}(\vec{x}) - \tilde{q}(\vec{x})\right| = S \left| \sigma\left(\frac{\tilde{q}(\vec{x})}{S}\right) - \frac{\tilde{q}(\vec{x})}{S}\right| \leq S \left(\frac{R}{S}\right)^3 = \delta \, \epsilon_{\Lambda}(f) \, .
  \end{align*}
  The $k > 0$ proof introduces no significantly new ideas but is more technical; see \cref{sec:sm-sobolev-theory}.
\end{proof}

The next result uses the $L^\infty$ (or $W^{k,\infty}$) approximability by SUPNs of the best-approximating polynomial to establish explicit error bounds of $f_{N, \Lambda}$ in the $W^{k,p}_\mu$ norm.

\begin{corollary} \label{cor:poly-approx}
    Under assumptions of \cref{prop:poly-approx}, for any $\delta > 0$, there is a SUPN of the form \cref{eq:ansatz_multidim} with $P = |\Lambda| + 1$ parameters such that 
    \begin{align}\label{eq:SUPN-polynomial}
        \left\|f - f_{N,\Lambda}\right\|_{W^{k,p}_\mu} < \left\{\begin{array}{rl} 
            (1 + \delta) \epsilon_{\Lambda}(f), & \textrm{ if }~\epsilon_{\Lambda}(f) > 0 \\
            \delta, & \textrm{ if }~\epsilon_{\Lambda}(f) = 0
        \end{array}\right. \quad .
    \end{align}
\end{corollary}
\begin{proof}
    Use $\delta/2$ and the triangle inequality in~\cref{prop:poly-approx}.
\end{proof}

\begin{remark}
  The results \eqref{eq:qtilde} and \eqref{eq:supn-q} ensure arbitrary $\delta$-accuracy relative to $\epsilon_{\Lambda}(f)$ for fixed parameter count $P$ at the cost of unbounded coefficients scaling like $1/\sqrt{\delta}$. In contrast, one can fix a $\delta$ (and hence \textit{a priori} ensure bounded SUPN coefficients) and instead drive the parameter count $P \uparrow \infty$ to decrease $\epsilon_{\Lambda}(f)$ to zero. Such a procedure is possible whenever polynomials are dense in the appropriate Banach space (e.g., \cite{ernst_convergence_2012} for $L^2$-type spaces). The interplay between $P$ and $\epsilon_{\Lambda}(f)$ depends on the smoothness of $f$. For example, if $f \in W_{\mu}^{r,2}$ and $k < r$, then $\epsilon_{\Lambda}(f)$ has an algebraic decay rate in $P$ that depends on $r - k$ \cite{bernardi_spectral_1997,gottlieb_numerical_1987,hesthaven_spectral_2007-1,trefethen_spectral_2000}. We provide one incarnation of such a result later in \cref{thm:constructive-Linf}.
\end{remark}

In contrast to the generality of approximability in Sobolev spaces we have shown above, if we specialize to $k = 0$ and $p = 2$, we can construct the near-optimal SUPN weights under the $L^2_\mu$ norm. This can be done by explicitly using the argumentation of \cref{proof:poly-approx}, even if $\mu$ is not the Chebyshev measure.

\begin{theorem}[Constructive SUPN in $L^2_\mu$ norm] \label{thm:constructive-L2}
    The SUPN guaranteed under \Cref{prop:poly-approx} with $k=0$ and  $p=2$ is given by
    \begin{align*}
        a_{1,\vec{m}} &= \frac{\tilde{\alpha}_{\vec{m}}}{S}, & c_1 &= S, & S&\coloneqq \begin{cases}
            \sqrt{\frac{R^3}{\delta}} & \quad \text{if } \epsilon_{\Lambda}(f) = 0 \\
            \sqrt{\frac{R^3}{\delta \, \epsilon_{\Lambda}(f)}} & \quad \text{if } \epsilon_{\Lambda}(f) > 0
        \end{cases}
    \end{align*}
    where $\{ \tilde{\alpha}_{\vec{m}} \}_{\vec{m} \in \Lambda}$ satisfy
    $$
    \sum_{\vec{m} \in \Lambda} \alpha_{\vec{m}} \phi_{\vec{m}} = \sum_{\vec{m} \in \Lambda} \tilde{\alpha}_{\vec{m}} T_{\vec{m}}
    $$
    and also
    \begin{align*}
        \alpha_{\vec{m}} &= \ip{\phi_{\vec{m}}}{f}_{L^2_\mu}, & R &= \sum_{\vec{m} \in \Lambda} |\tilde{\alpha}_{\vec{m}}|, \\
        \ip{\phi_{\vec{m}}}{\phi_{\vec{n}}}_{L^2_\mu} &= \delta_{\vec{m}, \vec{n}} \, , & \phi_{\vec{m}} &\in \P_{\Lambda} \ .
    \end{align*}
    Furthermore, the error $\epsilon_{\Lambda}(f)$ can be computed via
    \begin{equation}
        \epsilon_{\Lambda}(f) = \sqrt{\| f \|^2_{L^2_\mu} - \sum_{\vec{m} \in \Lambda} | \alpha_{\vec{m}} |^2}
    \end{equation}
\end{theorem}

\begin{proof}
Apply \cref{prop:poly-approx} with the optimal $L^2_\mu$ projection coefficients. 
\end{proof}

For the $L^\infty$ construction, we specialize the measure to the normalized
Chebyshev measure
\[
\mathrm{d}\mu_C(x)
=
\frac{\mathrm{d}x}{\pi\sqrt{1-x^2}},
\qquad x\in[-1,1].
\]
The Chebyshev polynomials of the first kind are orthogonal with respect to this
measure, so projection in $L^2_{\mu_C}$ provides explicit Chebyshev coefficients.
This projection is also nearly optimal in the $L^\infty$ norm~\cite{geddes1978near}.
Consequently, the constructive $L^2_\mu$ parameters from
\cref{thm:constructive-L2} can be specialized to $\mu=\mu_C$ to obtain the
$L^\infty$ construction below. The normalization makes $\mu_C$ a probability
measure, consistent with the assumptions of \cref{prop:poly-approx}, and does
not affect the projection coefficients because the normalization factor appears
in both the numerator and denominator.

\begin{theorem}[Constructive SUPN in $L^\infty$ norm and convergence rate] \label{thm:constructive-Linf}
    Let $f \in C(\Omega)$ with $\Omega \subset \R$ compact and $\delta >0$ be arbitrary. 
    
    \begin{enumerate}
        \item[(a)] We can construct a SUPN \cref{eq:ansatz} with $P=M+1$ parameters satisfying
        \begin{equation*}
            \| f - f_{1, M} \|_{L^\infty} < \left(  4 + \frac{4}{\pi^2}\log (M+1) \right) \inf_{q \in \P_M} \| f - q \|_{L^\infty} + \delta
        \end{equation*}
        using weights from \cref{thm:constructive-L2} with
        $S:=\sqrt{R^3/\delta}$ and $\mu=\mu_C$.
        \item[(b)] If $f \in C^k(\Omega)$ and $f^{(k)}$ is $\alpha$-H\"older continuous with $0 < \alpha \leq 1$, then there exists $\beta > 0$ such that
        \begin{equation*}
            \| f - f_{1, M} \|_{L^\infty} < \frac{\beta \log(M+1)}{M^{k+\alpha}} + \delta \, .
        \end{equation*}
    \end{enumerate}
\end{theorem}

\begin{proof}
    As shown in~\citet{rivlin1990chebyshev}, the Chebyshev series 
    \begin{align*}
    \tilde{f} &= \sum_{m \in [M]_0} \alpha_m T_m(x),
    &
    \alpha_m &= \frac{\ip{T_m}{f}_{L^2_{\mu_C}}}
    {\ip{T_m}{T_m}_{L^2_{\mu_C}}}.
    \end{align*}
    is nearly optimal as a degree-$M$ minimax polynomial, satisfying
    \begin{equation*}
        \| f - \tilde{f} \|_{L^\infty} < \left(  4 + \frac{4}{\pi^2}\log (M+1) \right) \inf_{q \in \P_M} \| f - q \|_{L^\infty} \, .
    \end{equation*}
    Using $S \coloneqq \sqrt{R^3 / \delta}$ and the weights from \cref{thm:constructive-L2} within \cref{proof:poly-approx} yields
    \begin{equation*}
        \| \tilde{f} - f_{1, M} \|_{L^\infty} < \delta ,
    \end{equation*}
    which completes the proof of part (a). Applying part (a) with Jackson's inequality (\cref{thm:jackson}) yields part (b).
\end{proof}

    \section{Second-order optimizer} \label{sec:optimizer}

As discussed in \cref{sssec:training-error}, best-approximation error can be obscured by insufficient optimization. We therefore use second-order optimization for SUPNs, DNNs, and KANs so that the numerical comparisons more closely reflect approximation capability rather than premature termination of training. Second-order methods have previously been studied for training neural networks; see \citet{xu2020second}. In our experiments, the training data is fixed, so gradients and Hessian information are deterministic rather than stochastic.

The resulting optimization problems are generally nonconvex, and their Hessians need not be positive definite. We therefore use a trust-region method, which approximately minimizes a local quadratic model subject to a step-size constraint,
\begin{align}
\begin{rcases}
    \displaystyle \min_{\vec{s} \in \R^P}
    & \quad
    (\nabla_{\vec{\theta}} \loss(\vec{\theta}_n))^\top \vec{s}
    + \frac{1}{2}\vec{s}^\top \vec{H}\vec{s}
    \\
    \text{s.t.}
    & \quad \|\vec{s}\| < \Delta_n
\end{rcases},
\label{eq:tr_subproblem}
\end{align}
where $\vec{H}$ is the Hessian $\nabla_{\vec{\theta}}^2 \loss(\vec{\theta}_n)$ or an approximation thereof and $\Delta_n$ is the trust-region radius. Standard details of trust-region methods can be found in \citet{nocedal2006numerical} and \citet{conn2000trust}.

For our experiments, the trust-region subproblem is solved using the Steihaug--Toint conjugate-gradient (CG) method \cite{Toint1981TowardsAE, steihaug1983conjugate}. We use a limited-memory BFGS approximation~\cite{liu1989limited} as a preconditioner and implement the optimization with the Rapid Optimization Library~\cite{javeed2022get}. The method does not require explicitly forming or inverting the Hessian; it requires only Hessian--vector products, which are computed using automatic differentiation.

\begin{figure}[htbp!]
    \centering
    \includegraphics[width=0.5\textwidth]{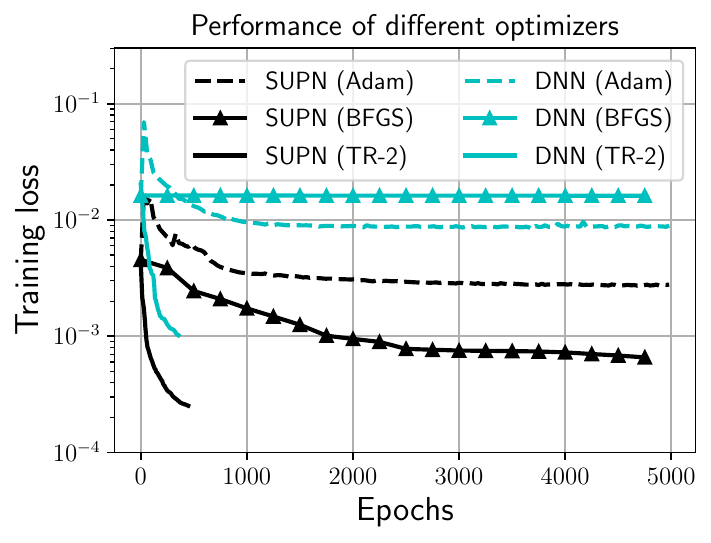}
    \caption{Training losses with Adam, BFGS, and a second-order trust-region method.}
    \label{fig:adam-comparison}
\end{figure}

\begin{example2}
    In \Cref{fig:adam-comparison}, we compare Adam~\cite{kingma2017adam},
    BFGS~\cite{liu1989limited}, and the second-order trust-region method after
    an initial 1000 epochs of Adam burn-in (not shown) for the target
    $f(x)=|x|$. Adam and BFGS use a learning rate of $10^{-2}$.
    Both DNNs and SUPNs attain lower training loss in substantially fewer
    epochs with Newton--CG than with Adam or BFGS, motivating its use in
    the best-approximation experiments.
\end{example2}

    \section{Numerical experiments} \label{sec:numerics}

We present several experiments in one, two, and ten dimensions to demonstrate the utility of SUPNs. These experiments are designed to show how SUPNs, DNNs, KANs, and polynomial projection perform on functions of varying regularity.\footnote{As discussed previously, Chebyshev KANs replace B-splines with Chebyshev polynomials. However, to maintain a tractable number of numerical experiments, we consider only the classic formulation of KANs with third-order B-splines and not Chebyshev KANs.} Importantly, they demonstrate the best-approximation error $\epsilon(\hat{\theta})$, which is free from error arising from insufficient sampling or premature stopping of the optimization method. Across these experiments, SUPNs are consistently competitive with DNNs, KANs, and polynomial projection, and often achieve the lowest error at a given parameter count. Furthermore, every method {\em except SUPNs} is the worst-performing method on at least one target function. These experiments are intended to characterize SUPN performance across target functions with different dimensionality, regularity, and structural properties; they do not imply uniform superiority over a probability distribution of arbitrary functions.

We briefly summarize the commonalities among all examples. Every DNN uses tanh activations in each layer except the output layer. KANs use a grid size of $5$, a spline order of $3$, and the SiLU activation function, in keeping with~\citet{liu2025kan}. The appendix contains specific values for $N$ and $M$ in each example. All experiments were performed on Nvidia A100 GPUs, except the 10D case, which used H100 GPUs due to memory requirements. The loss functions for neural networks are highly nonconvex in general, so we use a large number of allowable optimization iterations. Training consisted of a burn-in period of 5000 epochs with Adam (learning rate $10^{-3}$), followed by trust-region Newton--CG. The trust-region algorithm has a maximum of 1000 Newton steps, a gradient tolerance of $10^{-6}$ and a step tolerance of $5 \times 10^{-5}$. At each optimization step, CG has an absolute tolerance of $10^{-4}$, a relative tolerance of $10^{-2}$, and a maximum of 500 iterations. Finally, five independent realizations of initial network weights are performed for each experiment. Accuracy on a validation set is monitored during training, and the reported accuracy is the test-set accuracy at the minimum validation-set error, averaged over all runs. All data are noise-free since adding noise would create an artificial floor on the approximation error of all methods we consider here.

\subsection{One dimension}

We first compare best-approximation error and training robustness between SUPNs, DNNs, KANs, and polynomial projection on a suite of standard test problems. Then, we establish sampling requirements of SUPNs at models corresponding to low, medium, and high best-approximation accuracy. Next, we demonstrate that the order of convergence of SUPNs on the Runge function is independent of its bandwidth. Finally, we show that SUPNs can fit a highly oscillatory sine function from~\citet[Fig. 5]{ABUEIDDA2025117699} with an extended domain.

\subsubsection{Target functions}

\begin{figure}[htbp!]
    \centering
    \includegraphics[width=\textwidth]{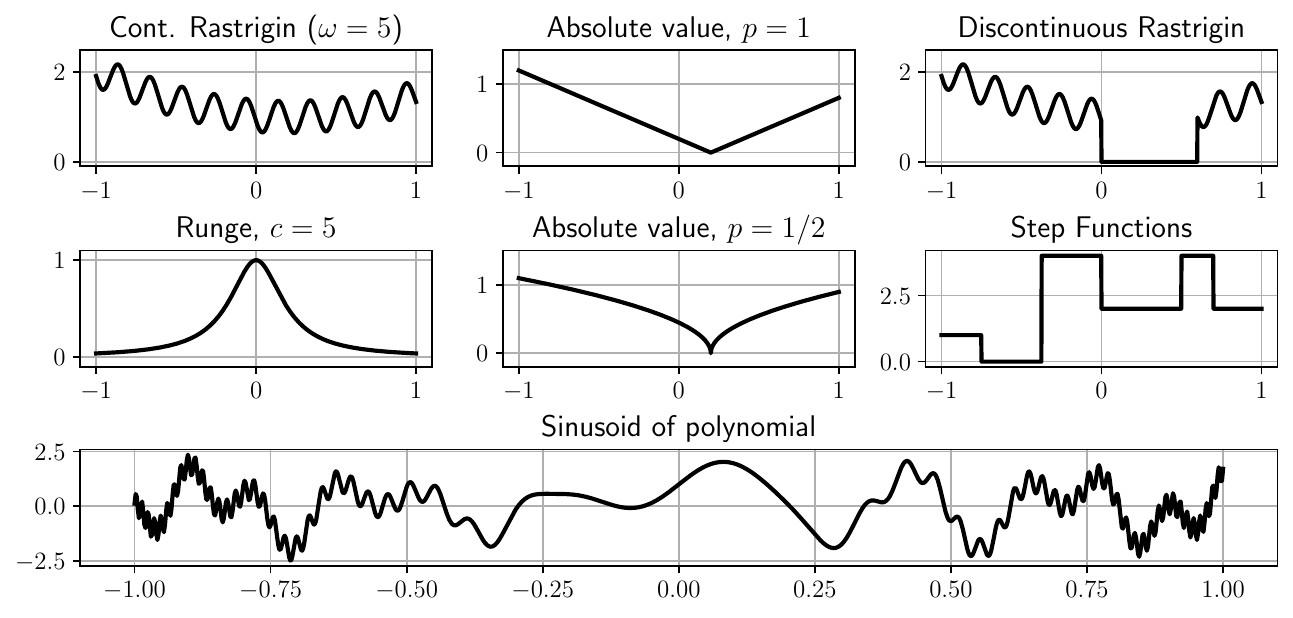}
    \caption{1D target functions for approximation.}
    \label{fig:1d-target-functions}
\end{figure}

We consider seven 1D functions $f: [-1, 1] \to \R$, shown in~\Cref{fig:1d-target-functions}. These functions are defined below.

\begin{enumerate}
    \item Continuous Rastrigin:
    \begin{equation}
    f_1(x; \omega) = 2(x-0.2)^2 - \frac{1}{2.77} \cos(2\pi\omega x - 1.22) + 1
    \label{eq:rastrigin}
    \end{equation}
    \item Discontinuous Rastrigin:
    \begin{equation*}
    f_2(x) = \begin{cases}
        0, & 0 \leq x \leq 0.6 \\
        f_1(x; \omega=5), & \mbox{otherwise}
    \end{cases}
    \end{equation*}
    \item Absolute value of order $p \in (0, 1]$:
    \begin{equation*}
    f_3(x; p) = |x - 0.2|^p
    \end{equation*}
    \item Linear combination of step functions:
    \begin{equation*}
    f_4(x) = \begin{cases}
        1, & -1 \leq x < -0.75 \\
        0, & -0.75 \leq x < -0.375 \\
        4, & -0.375 \leq x < 0~~\text{or}~~0.5 \leq x < 0.7\\
        2, & 0 \leq x < 0.5~~\text{or}~~0.7 \leq x \leq 1 \\
    \end{cases}
    \end{equation*}
    \item Runge function ($c \geq 1$):
    \begin{equation}
        f_5(x; c) = \frac{1}{1+(cx)^2}
        \label{eq:runge}
    \end{equation}
    \item Sinusoid of  polynomial:
    \begin{equation*}
    f_6(x) = \sin(2\pi^2x)+\cos(\pi^3x^2)+\cos(\pi^4x^3)\sin(\pi^4x^3)
    \end{equation*}
\end{enumerate}

\subsubsection{Best-approximation error} \label{ssec:best-approx-1d}

\begin{figure}
    \centering
    \includegraphics[width=\textwidth]{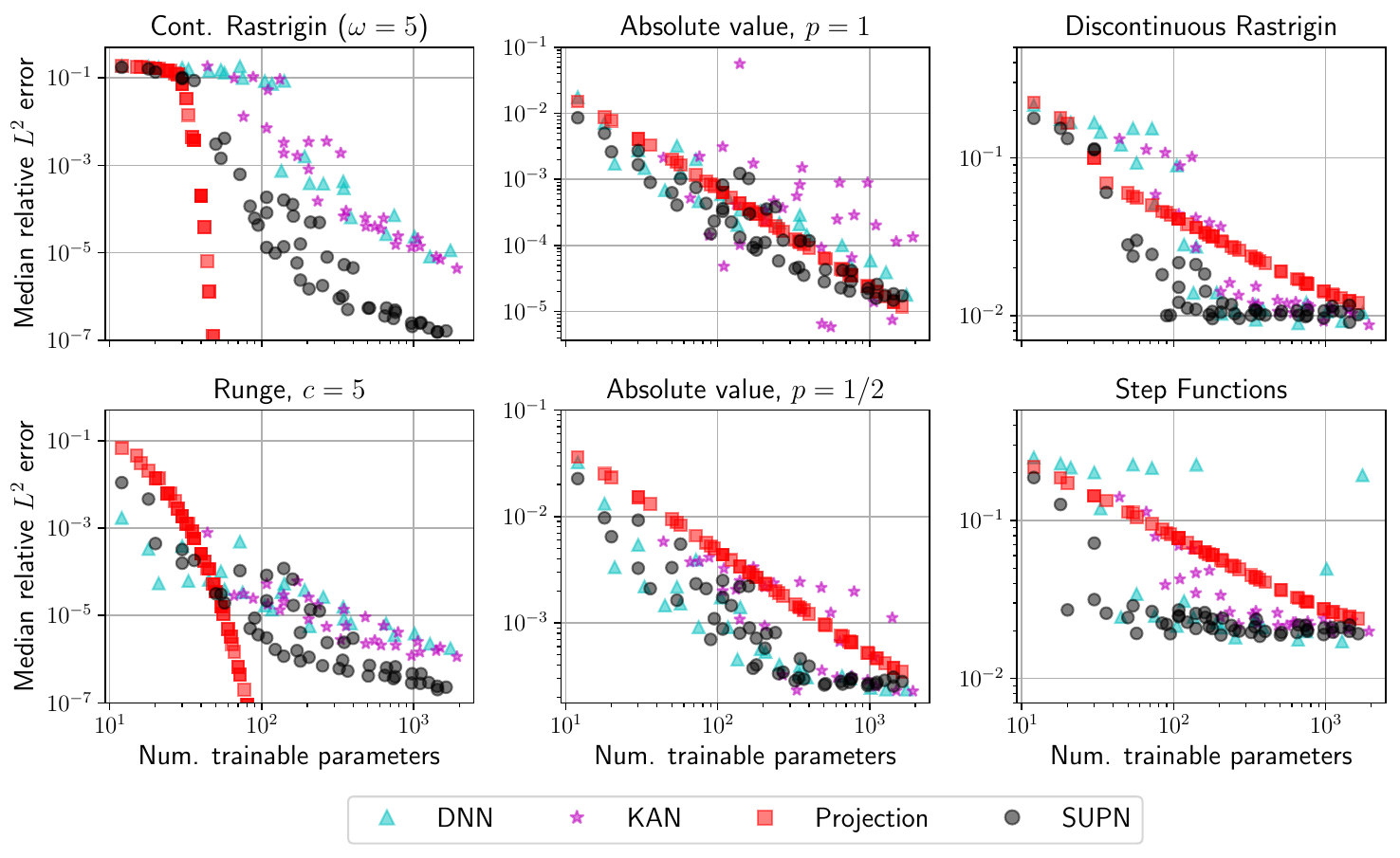}
    \caption{Relative $L^{2}$ errors vs. trainable parameters using SUPN, DNN, KAN, and polynomial projection for $f_k$, $k \in [5]$.}
    \label{fig:best-approx-1d}
\end{figure}

\begin{figure}
    \centering
    \includegraphics[width=\textwidth]{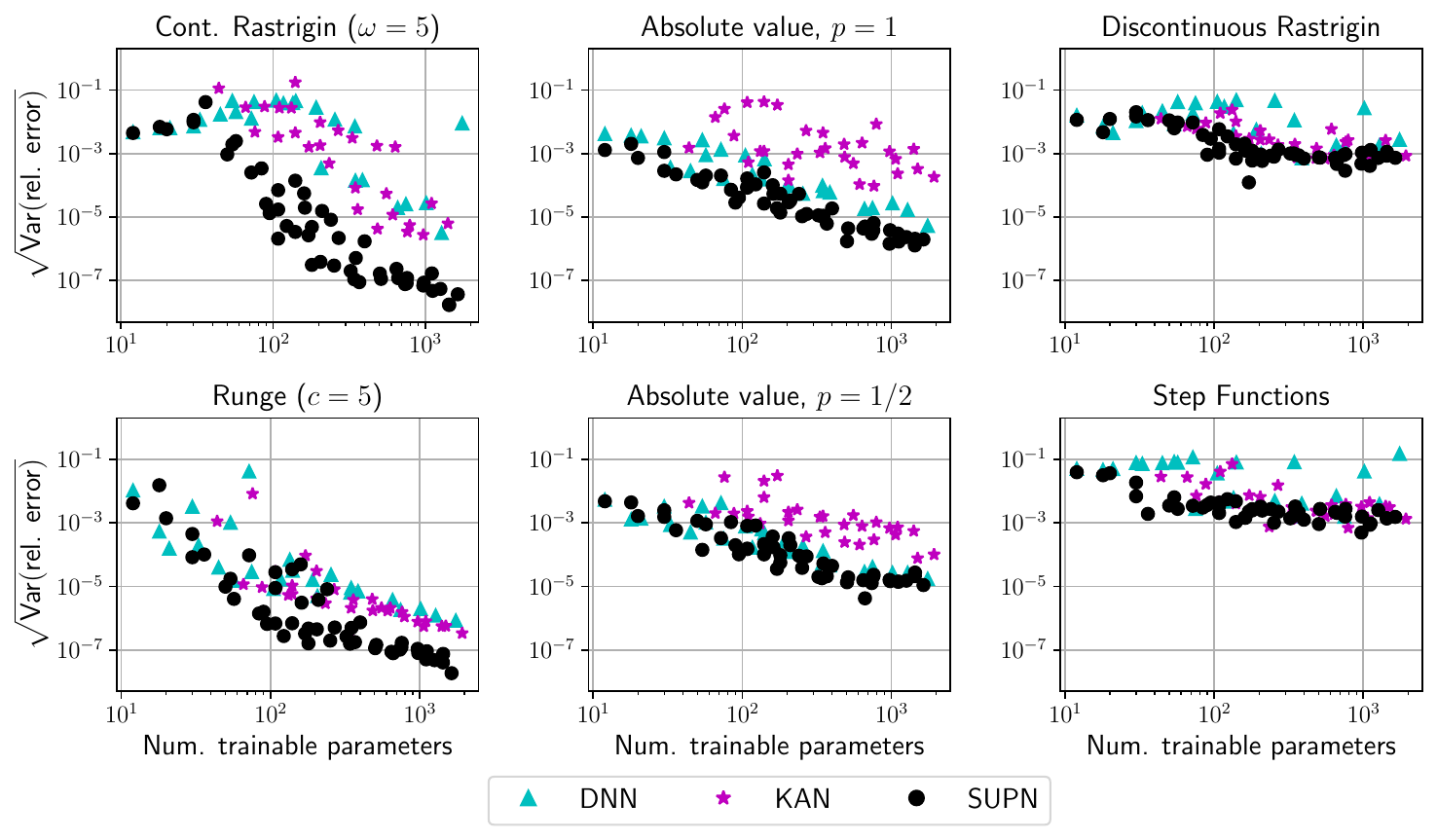}
    \caption{Standard deviation of relative error over 5 realizations of initial network weights.}
    \label{fig:training-robustness}
\end{figure}

We first examine the error of a local minimizer of~\cref{eq:best_surrogate}. We use a 2000-point Gauss--Legendre quadrature rule in the loss function~\cref{eq:loss}. The validation set consists of 3001 equidistant points on $[-1, 1]$. The test set consists of 17001 equidistant points on $[-1, 1]$, and we report the test error at the minimum validation error encountered during training. We vary $N$, $M$, width and depth to create networks with $P \leq 2000$; exact details are in the appendix (\cref{tab:sm-arch-1d}). We perform five initializations of network weights from the Kaiming uniform distribution~\cite{he2015delving}.

In~\Cref{fig:best-approx-1d}, we demonstrate that SUPNs are able to attain a given error tolerance with significantly fewer parameters than DNNs or KANs, often by an order of magnitude. For the continuously differentiable targets, i.e. continuous Rastrigin and Runge, projection outperforms all the other methods given sufficiently many trainable parameters, but SUPNs outperform DNNs and KANs. For $f(x)=|x|$, SUPNs have a very slight advantage over DNNs and projection when $10^2 \leq P \leq 10^3$. For $f(x) = |x|^{1/2}$, SUPNs and DNNs track each other very closely, outperforming polynomial projection; the infinite derivative at $x=0$ causes stagnation around $P \approx 500$. For discontinuous targets, SUPNs have the fewest parameters at the lowest error threshold. Similar trends hold for the $L^\infty$ norm (except for discontinuous targets, where all methods fail); for brevity, further details are in the appendix (\Cref{sec:sm-1D}).

\Cref{fig:training-robustness} shows the standard deviation of relative error over five realizations of initial network weights. The standard deviation decreases as the median relative error decreases, except in a handful of very small networks, which give reliably high error. Comparing \Cref{fig:best-approx-1d} and \Cref{fig:training-robustness} at a fixed number of trainable parameters shows that SUPNs generally exhibit lower variability than DNNs and KANs for a given parameter count. The caveat is $f(x) = |x|^{1/2}$, where DNNs and SUPNs are {\em equally} robust.

\subsubsection{Effect of hyperparameters}
\begin{figure}
    \centering
    \includegraphics[width=0.9\textwidth]{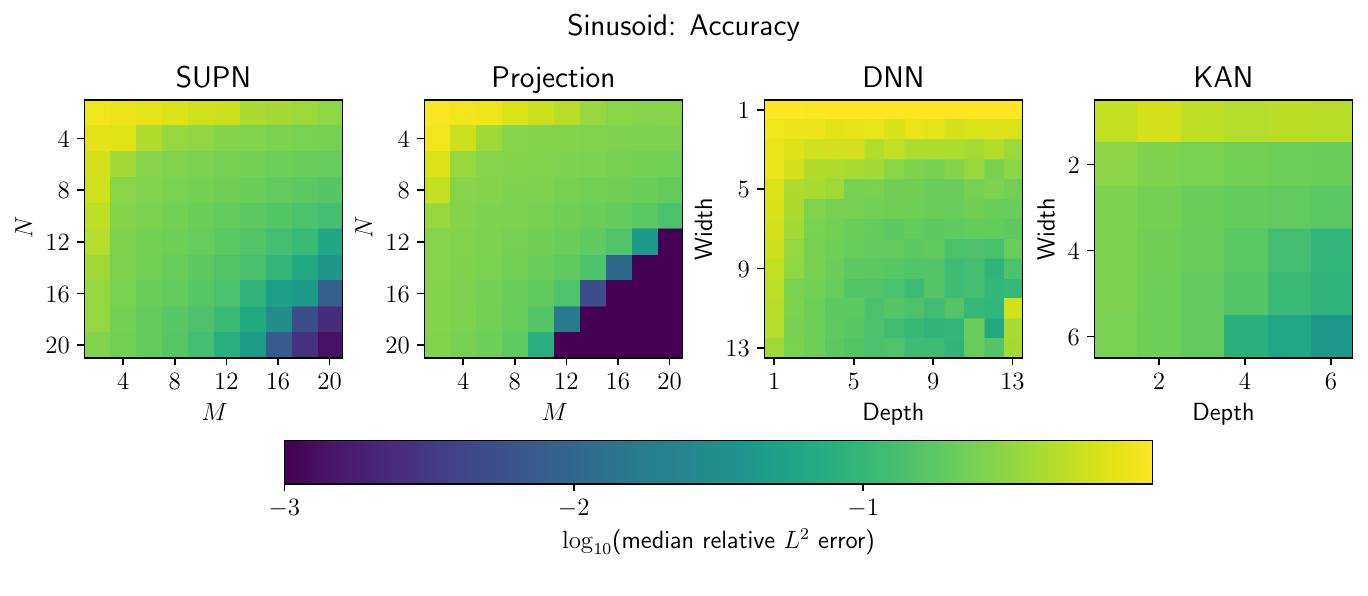}
    \includegraphics[width=0.9\textwidth]{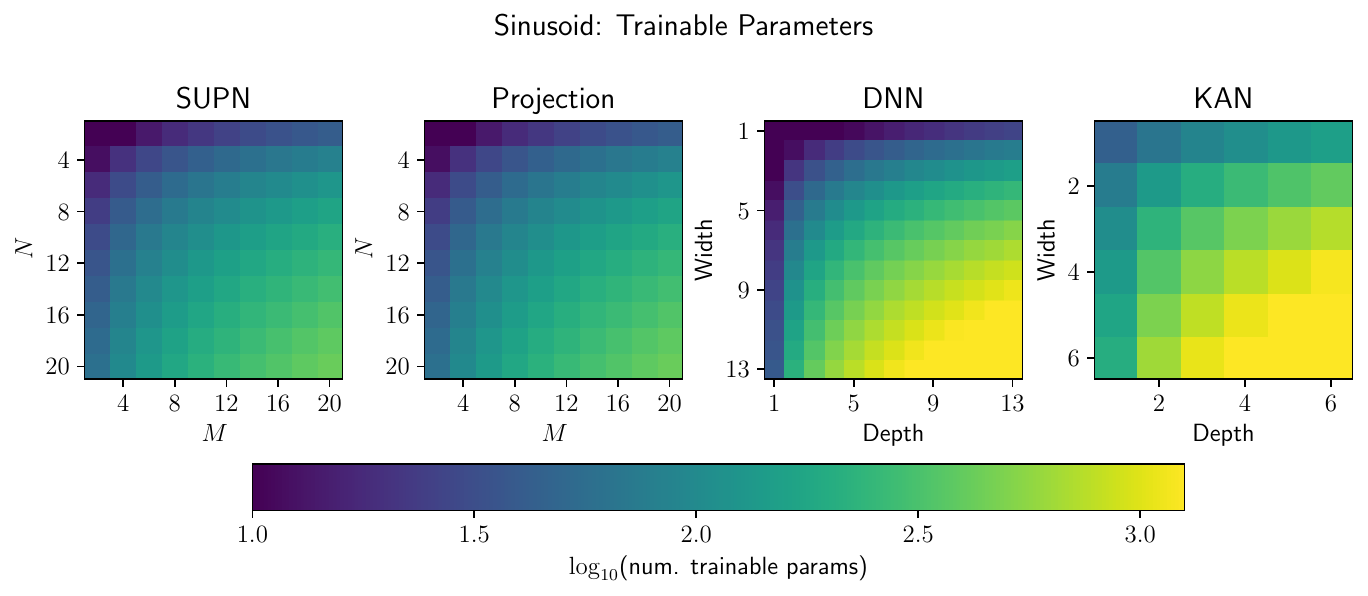}
    \includegraphics[width=0.9\textwidth]{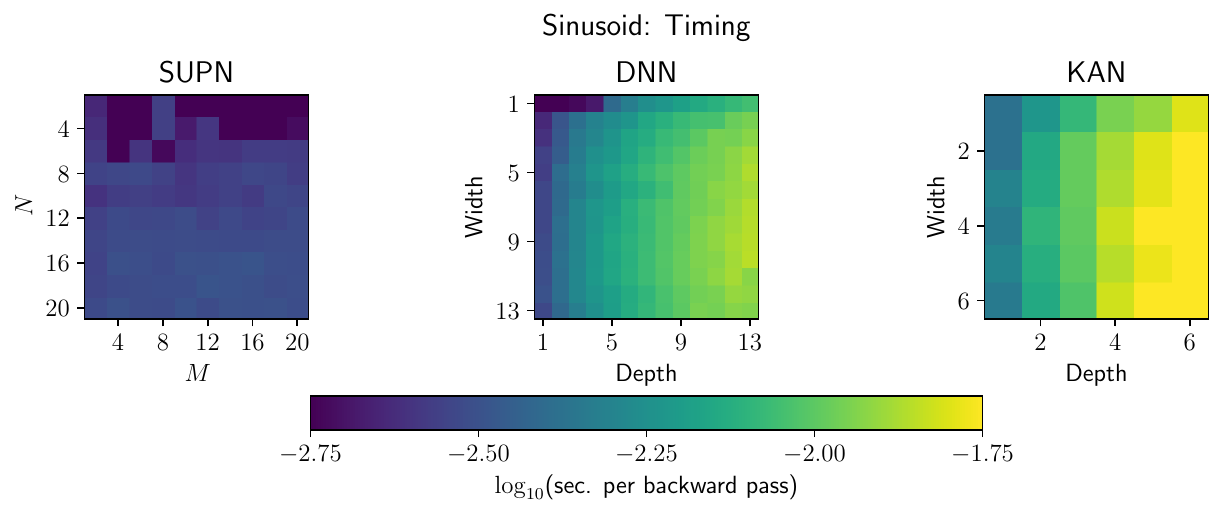}
    \caption{Top: Heat map of test errors for $f_6(x)$ over the hyperparameter sweep. Middle: Corresponding numbers of trainable parameters. Bottom: Time required to compute one backward pass for SUPNs, DNNs, and KANs, averaged over 5000 Adam and 1000 Newton--CG iterations for each architecture. Polynomial projection is omitted because it does not require iterative training, it also takes much less time than the neural networks to perform function approximation. Direct numerical comparisons are in \Cref{fig:best-approx-1d}. Compared to DNNs and KANs, SUPNs display continuous accuracy improvement in $(N,M)$ along with far lower parameter counts and relative errors.}
    \label{fig:hyperparameters}
\end{figure}

\Cref{fig:hyperparameters} shows the best-approximation error, number of trainable parameters,
and computational cost across the SUPN, DNN, KAN, and polynomial-projection hyperparameter sweeps for the sinusoidal target, $f_6(x)$. The middle row shows
how $P$ varies with different choices of $N$ and $M$. The accuracy results indicate that overly narrow SUPNs (small $N$) can saturate the approximation capability of a degree-$M$ polynomial in practice, for example when the learned parameters differ from the constructive initialization in \cref{thm:constructive-L2}. Likewise, SUPNs with low polynomial degree $M$
require a larger number $N$ of basis functions to achieve comparable accuracy. Across the $(N,M)$ sweep, SUPNs generally improve continuously in accuracy as
either source of model complexity is increased, whereas the DNN and KAN results exhibit sharper transitions in accuracy with respect to width and depth.

The timing results show an additional distinction between the two SUPN complexity parameters. Increasing $M$ at fixed $N$ produces comparatively little growth in the cost of a backward pass, while increasing $N$ results in a clearer increase in computational cost. Thus, increasing polynomial complexity provides a relatively inexpensive direction for increasing SUPN
approximation capacity. This empirical behavior complements \cref{thm:fixed_N}, which establishes universal approximation at fixed finite width by increasing $M$, while the corresponding variable-width mechanism is given by \cref{thm:fixed_M}.

\begin{figure}
    \centering
    \includegraphics[width=0.9\textwidth]{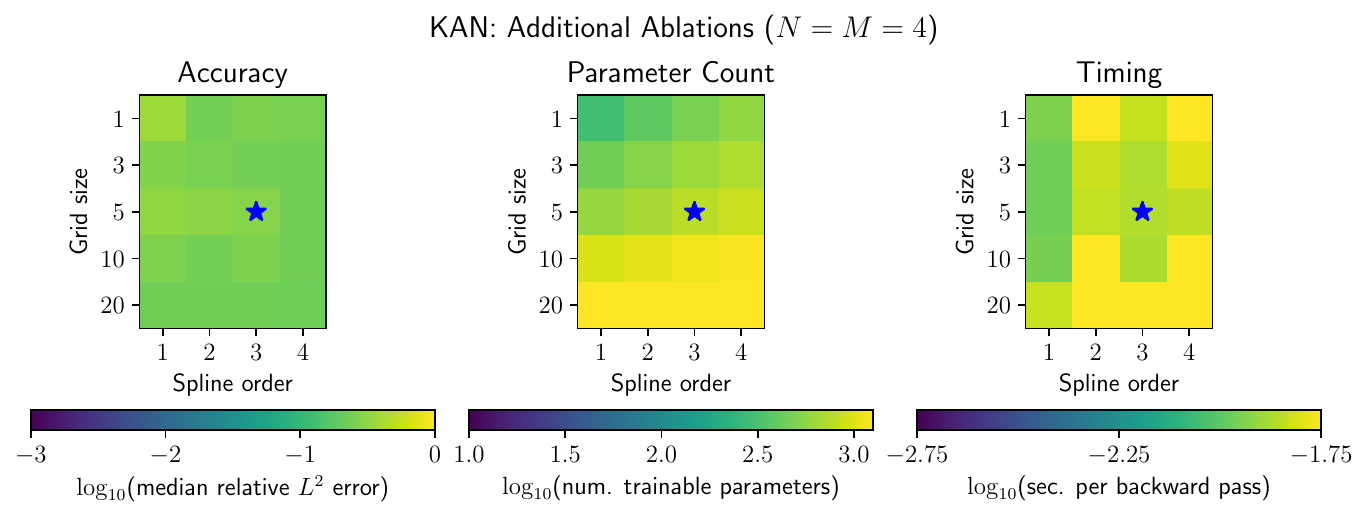}
    \caption{Ablation study for KANs over grid size $G$ and spline order for fixed width of $N=4$ and depth $M=4$. The colorbar extents are the same as in \Cref{fig:hyperparameters}. The blue star indicates the default choice of $G=5$ and third-order splines used in the other KAN results. There is no meaningful change in KAN accuracy over the parameter sweep for this problem.}
    \label{fig:kan_grid_order_ablation}
\end{figure}

By default, we consider B-spline KANs with grid size $G=5$ and third-order splines. As justification for this choice, \Cref{fig:kan_grid_order_ablation} shows a sweep over grid size and spline order for a fixed width and depth. Accuracy does not meaningfully change over the sweep.

\subsubsection{Sampling error}
\begin{figure}
    \centering
    \includegraphics[width=\textwidth]{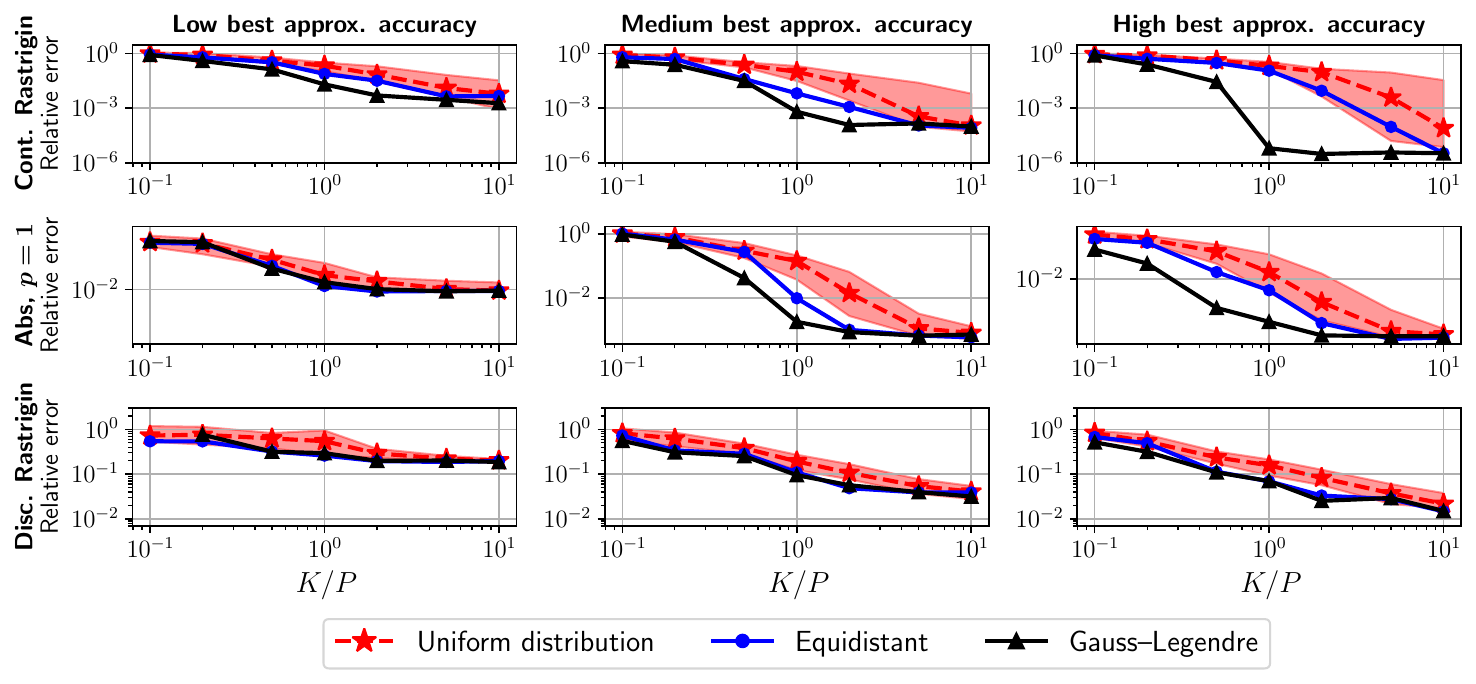}
    \caption{Median $L^2$ error (over network initialization) of finite-sampling experiments for different combinations of target function, best approximation accuracy, and sampling strategy. For the ``uniform'' results, the line is the median over 10 realizations of training data and 5 weight initializations, while the shaded region is the 10th to 90th percentile.}
    \label{fig:finite-sampling}
\end{figure}

Next, we empirically investigate sampling requirements for SUPNs. We vary the ratio between the number of training data $K$ and the number of SUPN parameters, $P = N(M+1)$. We examine models corresponding to qualitatively low, medium, and high best-approximation accuracy from \cref{fig:best-approx-1d}, using the sampling techniques discussed in~\cref{ssec:accuracy-estimation}. \Cref{fig:finite-sampling} indicates that Gauss--Legendre requires the fewest number of samples to achieve best-approximation error. The specific ratio appears to depend on the smoothness of the target function. We see up to $K=P$ for continuous Rastrigin, and $K=2P$ for absolute value and discontinuous Rastrigin.

\subsubsection{Convergence on Runge function}
\begin{figure}
    \centering
    \includegraphics[width=0.5\textwidth]{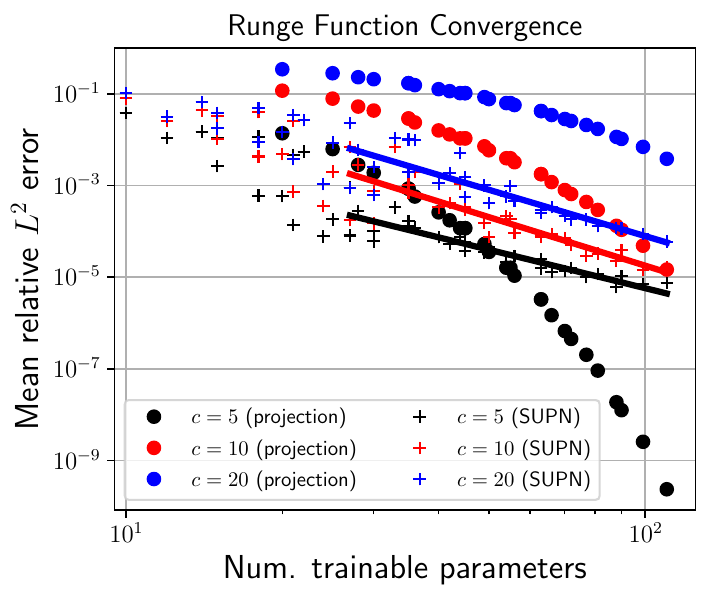}
    \caption{Convergence of SUPN and polynomial projection on Runge function for $c=5, 10, 20$. Lines of best fit are shown for SUPN.}
    \label{fig:runge}
\end{figure}

The Runge function \cref{eq:runge} is a famous example of a function that can be difficult to approximate with naive methods~\cite{atkinson1989introduction}. Consider approximating the parameterized Runge function with polynomial projection in the $L^2$ norm. It can be shown that the best approximation error~\cref{eq:best-approx-error} using the $P$-term Legendre series~\cref{eq:legendre-series} is bounded by
\begin{equation*}
    \epsilon(\vec{\surrbest{\theta}}) \lesssim \exp(-\beta P/c)
\end{equation*}
for some universal constant $\beta > 0$. See, e.g.,~\citet[Theorem 2.1]{wang2012convergence} or \citet[Theorem 8.2]{trefethen_approximation_2012}.  This bound gives spectral convergence, but the convergence rate decays with $c$. \Cref{fig:runge} shows spectral convergence for polynomial projection that degrades as $c$ increases. In contrast, SUPNs display a finite order of convergence whose observed slope is approximately unchanged across the values of $c$ considered, although the error magnitude increases with $c$.

\subsubsection{Sinusoid of polynomial}
\begin{figure}
    \centering
    \includegraphics[width=\textwidth]{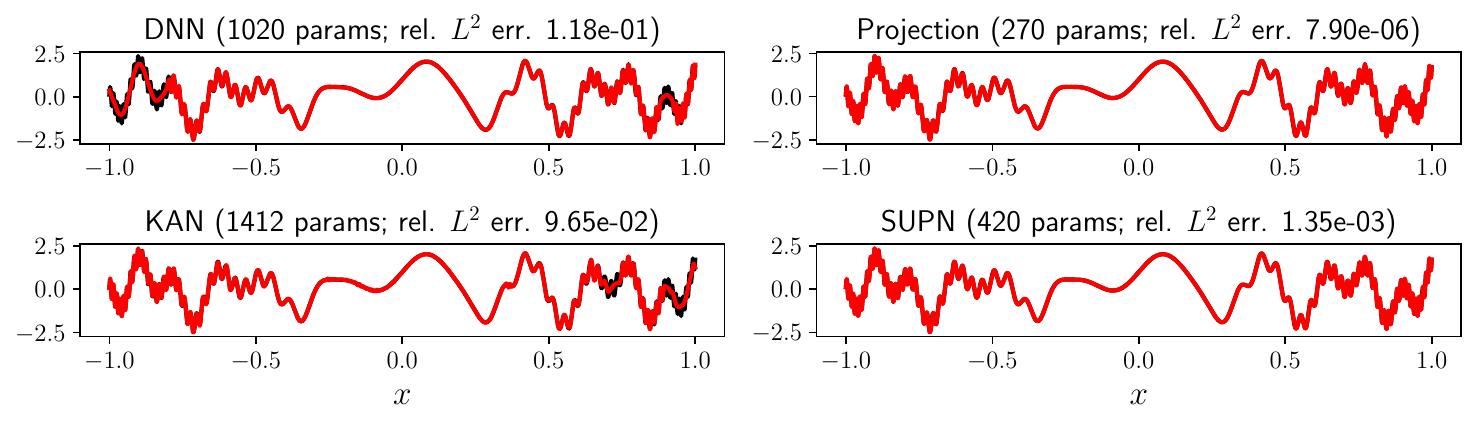}
    \caption{Surrogate predictions on sinusoid example corresponding to median-error model.}
    \label{fig:oscillatory-sine}
\end{figure}

This function uses an extended domain of the problem in \citet[Sec.~3.1]{ABUEIDDA2025117699}, in which the authors observe MLPs struggling to fit high-frequency components of the function. We display these results separately from the other functions because larger networks are required than for the experiments in \Cref{fig:best-approx-1d}. \Cref{fig:oscillatory-sine} demonstrates that SUPNs and polynomial projection have low error with a small number of parameters, while KANs and especially MLPs struggle at higher frequencies.

\subsubsection{Gaussian processes}

In addition to the target functions in \Cref{fig:1d-target-functions}, we also consider functions drawn from a Gaussian process (GP). This allows a wider range of target functions with varying smoothness to illustrate the expressive capability of SUPNs. We use a prior mean of zero and a Mat\'{e}rn covariance kernel $k(x, y)$ with smoothness parameter $\nu \in \{ 1/2, 3/2, \infty \}$:
\begin{alignat*}{2}
k_{\infty}(x, y; \ell) &= \exp\left( -\frac{(x-y)^2}{2\ell^2} \right), && \qquad \ell = 0.1, \\
k_{3/2}(x, y; \ell) &= \left( 1 + \sqrt{3}\frac{|x-y|}{\ell} \right) \exp \left( -\sqrt{3} \frac{|x-y|}{\ell} \right), && \qquad \ell = 0.1, \\
k_{1/2}(x, y; \ell) &= \exp\left( - \frac{|x-y|}{\ell} \right), && \qquad \ell = 10. \\
\end{alignat*} 
These kernels induce realizations of functions that are, respectively, $C^\infty$, $C^1$, and $C^0$.

\begin{figure}
    \centering
    \includegraphics[width=0.9\textwidth]{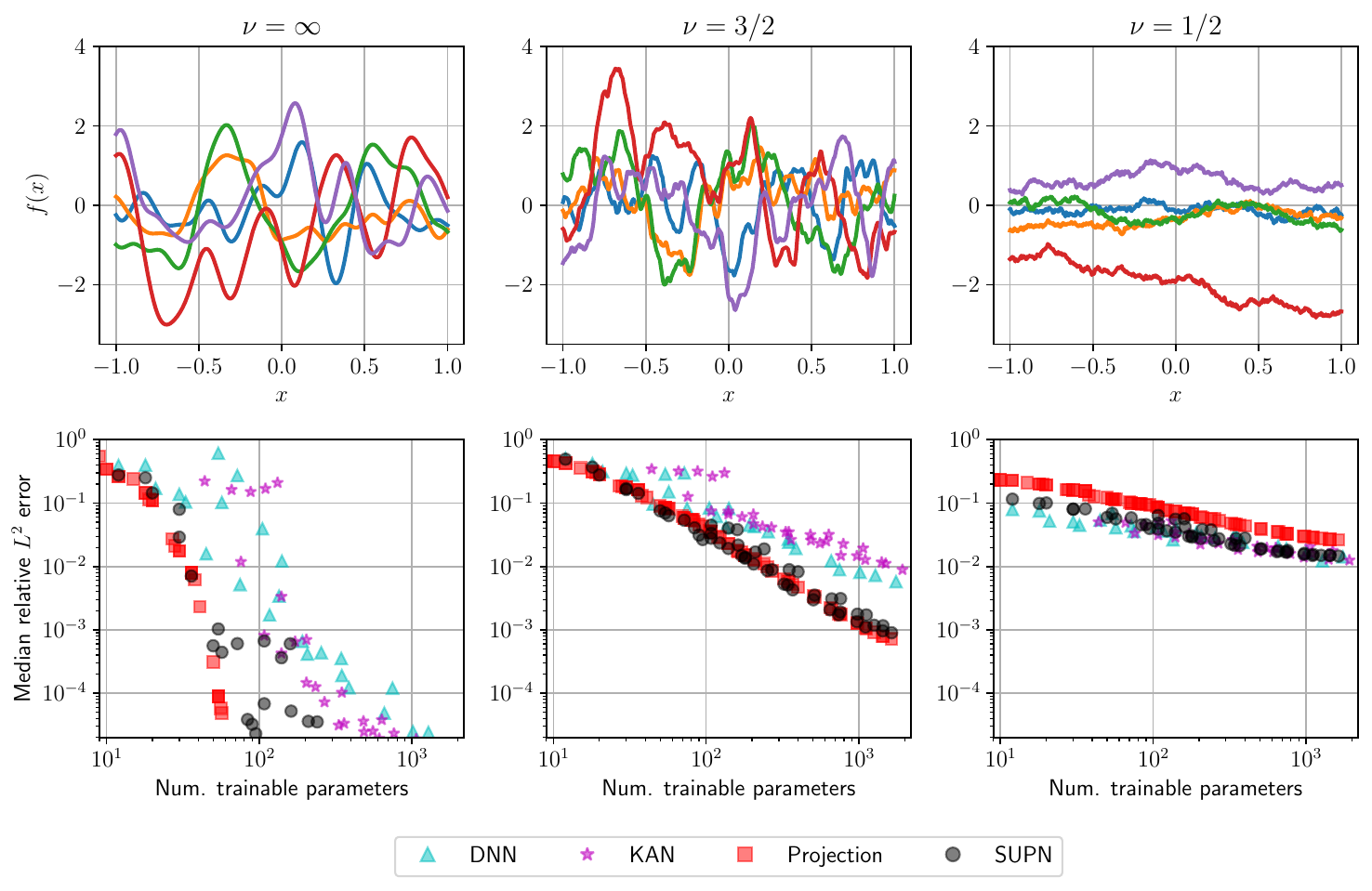}
    \caption{Top: GP realizations. Bottom: Median relative $L^2$ error over network initialization and GP draws.}
    \label{fig:GP}
\end{figure}

We apply the same methods as before to five target functions from each GP using a common length-scale $\ell = 0.1$ and show the results in \Cref{fig:GP}. We report the median relative $L^2$ error over both network initialization and GP draws. In these examples, SUPNs are never the worst-converging method, while DNNs, KANs, and polynomials sometimes are, depending on the smoothness of the target function. As expected, convergence is qualitatively similar to that of \Cref{fig:best-approx-1d}. On analytic functions ($\nu = \infty$), polynomials have the advantage, followed by SUPNs and then DNNs/KANs. On finitely differentiable functions ($\nu = 3/2$), polynomials and SUPNs converge almost identically, in keeping with \Cref{prop:poly-approx}, while DNNs/KANs are noticeably slower. On non-smooth functions ($\nu=1/2$), DNNs/KANs/SUPNs converge almost identically, while polynomials are noticeably slower.

\subsection{Two dimensions} \label{ssec:2d}
We now present best-approximation results for functions with two input dimensions. The results of a finite-sampling study are in the appendix (\cref{fig:sm-finite-sampling-supn-2d}) and qualitatively similar to the one-dimensional case.
For both SUPN and projection, we use the isotropic hyperbolic cross-section $\Lambda_{\text{HC}}$ \cref{eq:lambda-hc} and total-degree space $\Lambda_{\text{TD}}$ \cref{eq:lambda-td} for admissible polynomial degrees.

\begin{figure}
    \centering
    \includegraphics[width=\textwidth]{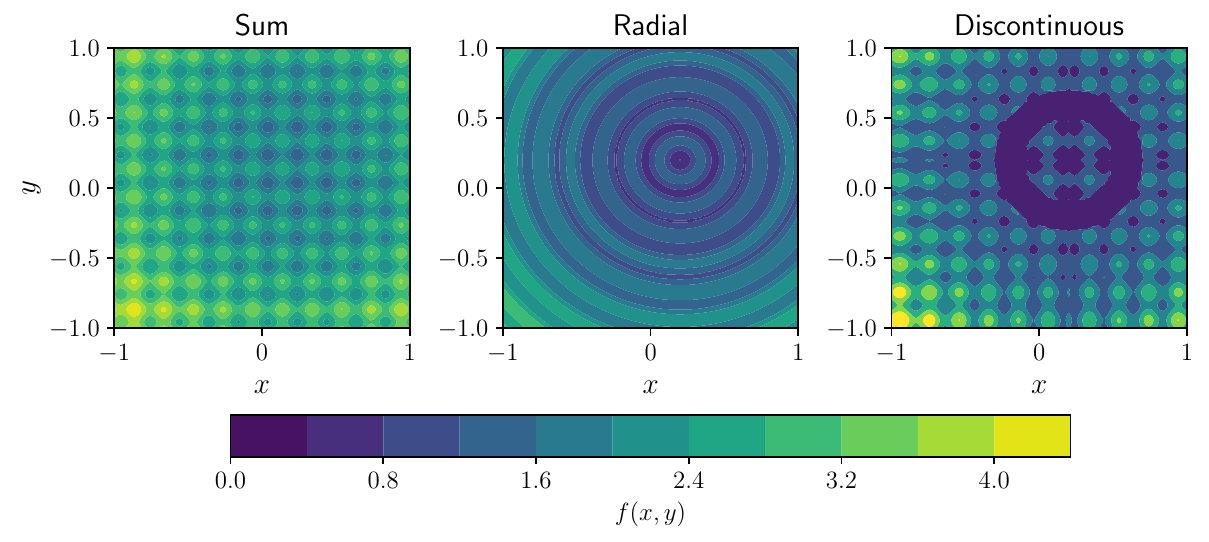}
    \caption{2D target functions for approximation.}
    \label{fig:2d-target-functions}
\end{figure}

\subsubsection{Target functions}
We consider three 2D functions $f: [-1,1]^2 \to \R$, based on the one-dimensional Rastrigin function~\cref{eq:rastrigin} with $\omega=5$. 
These target functions are shown in \Cref{fig:2d-target-functions} and are selected to have different amounts of regularity.

\begin{enumerate}
    \item Continuous Rastrigin (sum):
    \begin{align*}
    f_7(x,y) =  f_1(x; \omega=5) + f_1(y; \omega=5)
    \end{align*}
    \item Continuous Rastrigin (radial):
    \begin{align*}
    f_8(x,y) =  f_1(r; \omega=5), & \qquad 
    r = \sqrt{(x-0.2)^2+(y-0.2)^2}
    \end{align*}
    \item Discontinuous Rastrigin ($\omega=5$):
    \begin{equation*}
    f_9(x,y) =  \begin{cases}
    0,  & r \in [0.3, 0.5] \\
    f_1(|x-0.2|)f_1(|y-0.2|), &\text{otherwise}
    \end{cases}
    \end{equation*}
\end{enumerate}

\subsubsection{Best-approximation error}
In the loss function~\cref{eq:loss}, we use a tensor-product Gauss--Legendre rule with 200 nodes in each dimension. The validation set consists of a $130 \times 130$ grid of equispaced points, and the test set consists of a $450 \times 450$ grid of equispaced points. Due to the computational cost of a tensor-product grid, the validation and test sets are commensurate in size with the training set, unlike in \cref{ssec:best-approx-1d}.

\begin{figure}
    \centering
    \includegraphics[width=\textwidth]{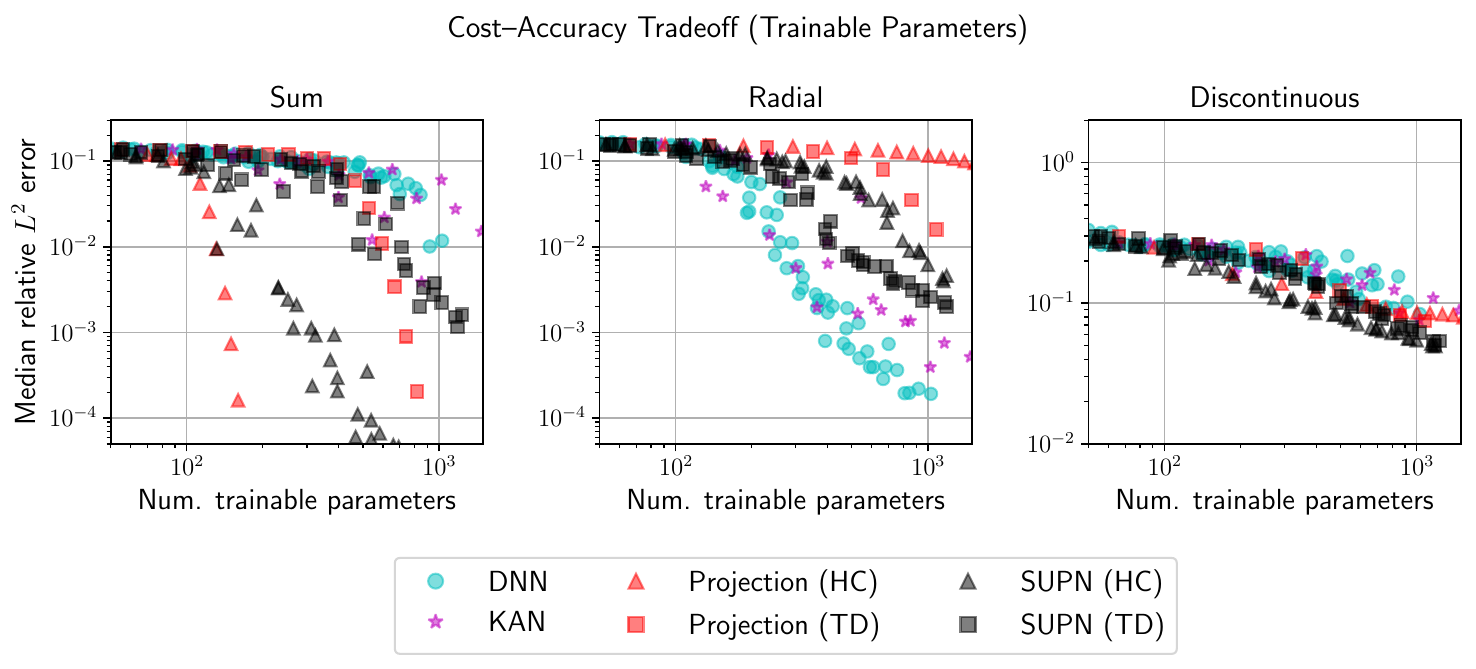}
    \caption{Median $L^2$ error versus number of trainable parameters $P$ for 2D examples ($f_k$, $k = 7, 8, 9$).}
    \label{fig:best-approx-2d}
\end{figure}

\Cref{fig:best-approx-2d} shows that SUPNs are a competitive method in 2D. For a sum of one-dimensional Rastrigin functions, SUPNs dramatically outperform DNNs and KANs. SUPNs with a properly chosen index set $\Lambda$ even outperform polynomial projection with a misspecified $\Lambda$. Interestingly, DNNs outperform all methods on the radial target, which we explain by the lack of tensor-product structure in $f_8$ causing difficulty for polynomial-based methods.  However, with a transform into polar coordinates $(r, \theta)$, the radial results would become virtually identical to the top left panel of~\cref{fig:best-approx-1d}. Moreover, the horizontal spread of KANs on the radial target is much wider than for SUPNs, so the performance of KANs is less predictable with respect to its width and depth. On the discontinuous target, SUPNs have the lowest error among all methods at a given parameter count. Robustness on these 2D examples is qualitatively similar to \cref{fig:training-robustness} and may be found in the Appendix (\cref{fig:sm-training-robustness-2d}).

\subsection{Ten dimensions}

We now demonstrate that SUPNs can scale into moderately large dimensions. We consider the anisotropic and nonlinear target function
\begin{align*}
f_{\text{aniso}}(\vec{x}) &= \exp(x_1-0.7)\sin(1.3x_2) + 0.2 \cos(2\pi x_3) + 0.01|x_4-0.27|x_5 \\
& \qquad + 0.1 |x_6|x_7 + 0.05 \exp(-(x_8-0.3)^2/16) + 0.1x_9 x_{10}
\end{align*}
where $\vec{x} \in [-1, 1]^{10}$. We use  $10^5$ samples from the Halton sequence for training. For validation, we use the next $2 \times 10^5$ Halton elements, and for testing, the {\em next} $2 \times 10^5$ Halton elements. As in \cref{ssec:2d}, all index sets are isotropic.

\begin{figure}
    \centering
    \includegraphics[width=0.7\textwidth]{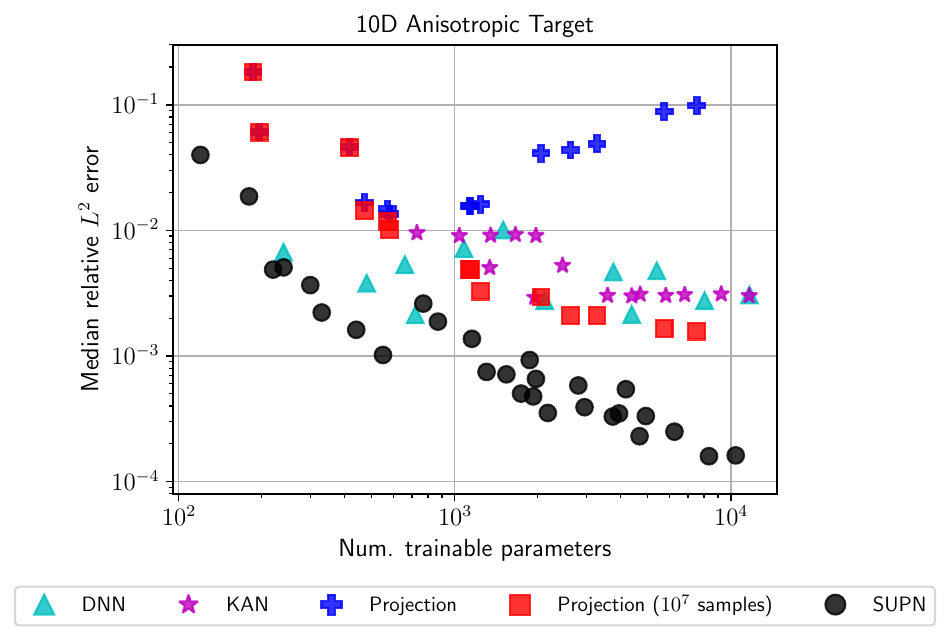}
    \caption{Median relative $L^2$ error on anisotropic target function with $D=10$ input dimensions.}
    \label{fig:best-approx-10d}
\end{figure}

\Cref{fig:best-approx-10d} displays the results. Asymptotically, SUPNs are an order of magnitude more accurate than both KANs and DNNs. To ensure we show best-approximation error rather than sampling error, we also show projection using $10^7$ Halton elements to compute the $L^2$ inner products. With only $10^5$ samples, there is significant quadrature error in the higher-order Legendre expansion coefficients ($P \gtrsim 10^3$).

    \section{Conclusion and future work} \label{sec:conclusion}

We have presented shallow universal polynomial networks, a parsimonious and provably convergent function approximator that combines a shallow network architecture with learnable multivariate polynomial pre-activations. Beyond universal approximation, we establish finite-complexity guarantees showing that SUPNs can approach the best approximation attainable in a prescribed polynomial space to arbitrary relative tolerance, together with explicit quasi-optimal parameter constructions. Our numerical experiments show that SUPNs frequently achieve lower approximation error than DNNs and KANs at comparable parameter counts across the synthetic one-, two-, and ten-dimensional target functions considered. Taken together, these results suggest that SUPNs provide a robust choice for function approximation when the structure and regularity of the target are not known \textit{a priori}, a setting that commonly arises in surrogate modeling and the PDE applications motivating our ongoing work.

Extensions of SUPNs to operator learning and physics-informed PDE solution are the subject of ongoing work. One direction uses SUPNs as the trunk network of DeepONets \cite{lu2021learning} while a second develops physics-informed \cite{karniadakis2021physics} SUPNs for building surrogates on real-world systems without data. Initial results in both settings show improvements over the corresponding standard architectures. These extensions introduce additional questions associated with operator approximation and physics-informed optimization that are outside the scope of the present study of data-driven function approximation. We also plan to investigate adaptive strategies for constructing $\Lambda$ so that SUPNs can better exploit anisotropy in higher-dimensional function-approximation problems.

    \subsubsection*{Broader Impact Statement}
    This work does not directly cause the negative impacts listed in the TMLR Ethics Guidelines.

    \subsubsection*{Acknowledgments}
    The authors gratefully acknowledge funding from Sandia National Laboratories' Laboratory Directed Research and Development program. Sandia National Laboratories is a multimission laboratory managed and operated by National Technology and Engineering Solutions of Sandia, LLC, a wholly owned subsidiary of Honeywell International Inc., for the U.S. Department of Energy’s National Nuclear Security Administration under contract DE-NA0003525. Any subjective views or opinions that might be expressed in the paper do not necessarily represent the views of the U.S. Department of Energy or the United States Government. The DOE will provide public access to results of federally sponsored research in accordance with the DOE Public Access Plan.

    \bibliography{refs}
    \bibliographystyle{tmlr}

    \appendix

\section{Symbols and Notation} \label{app:symbols}

\begin{table}[H]
	\centering
	\begin{tabular}[c]{ll||ll}
    \hline
    $D$ & Input dimension & $\mathcal{L}(\cdot)$ & Loss function \\
    $Q$ & Output dimension & $\vec{\theta}$ & Learnable parameters \\
    $P$ & Parameter dimension & $\vec{\surrbest{\theta}}$ & Best-approximation parameters \\
    $\Omega$ & Compact subset of $\R^D$ & $\vec{\surr{\theta}}$ & Empirically trained parameters \\
    $T_i$ & Chebyshev polynomial of deg. $i$ & $\sigma$ & Activation function  \\
    $L_i$ & Legendre polynomial of deg. $i$ & $B_{j,i}$ & B-spline of deg. $i$ on partition $j$ \\
    $K$ & Training set size & $\epsilon$ & Generalization error \\
    $[N]$ & $\{1,2, \dots, N\}$ & $[N]_0$ & $\{0, 1, \dots, N\}$ \\
    $\P_n$ & Polynomials of degree $n$ & & \\
    \hline \\
	\end{tabular}
    \caption{Symbols and notation.}
\end{table}

\section{Supplementary 1D results} \label{sec:sm-1D}

\subsection{$L^\infty$ best-approximation error}
\begin{figure}
    \centering
    \includegraphics[width=\textwidth]{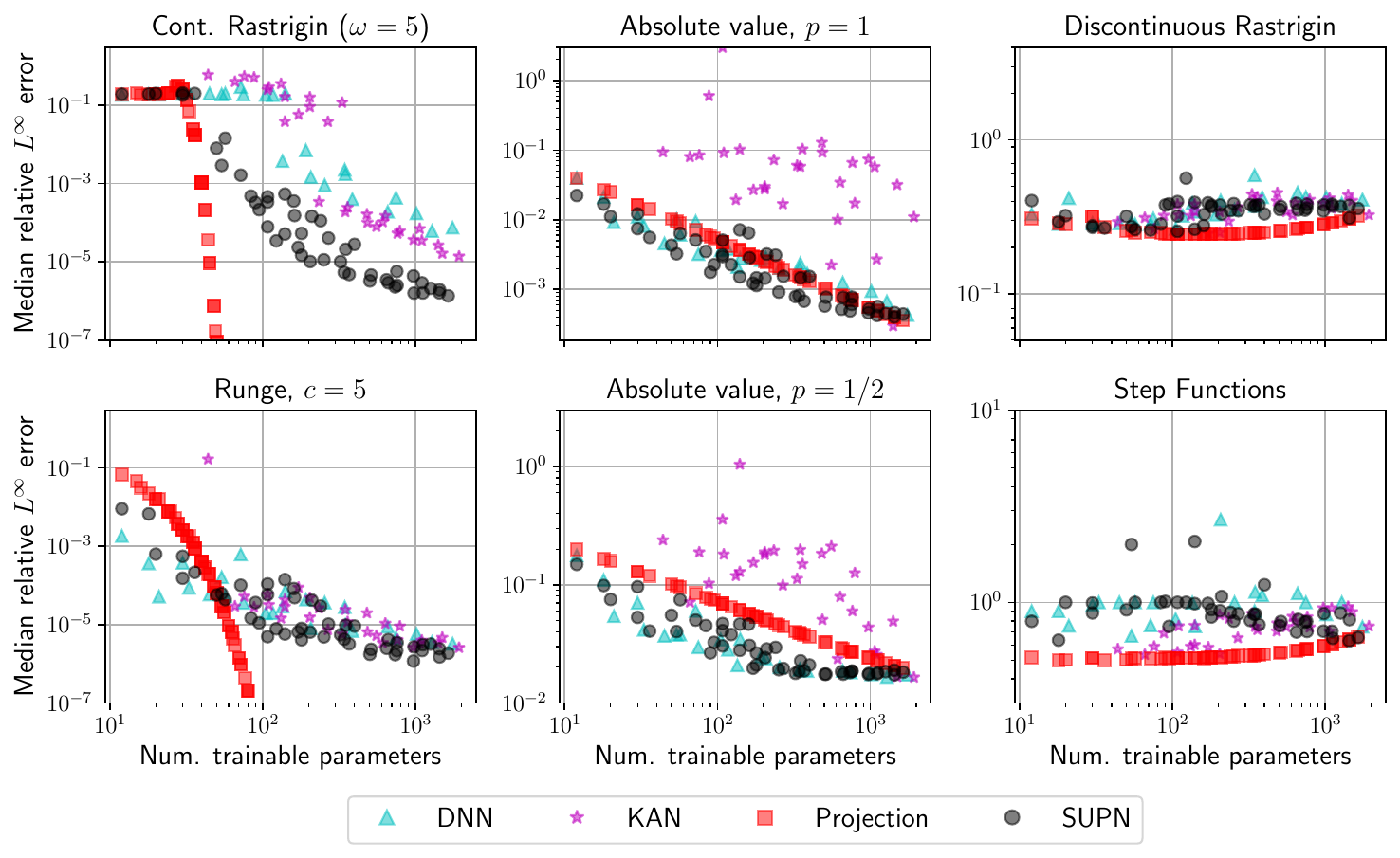}
    \caption{Relative $L^{\infty}$ errors vs. trainable parameters using SUPN, DNN, KAN, and polynomial projection for $f_k$, $k \in [6]$.}
    \label{fig:sm-best-approx-1d-L-inf}
\end{figure}

In \Cref{fig:sm-best-approx-1d-L-inf}, we observe similar convergence trends in learnable parameter count for $L^{\infty}$ as we do in $L^2$ for continuous function approximation cases. The exceptions are that KANs perform noticeably worse on $f(x) = |x|^p$, and SUPNs are now evenly matched on the Runge function. For discontinuous cases, all methods suffer in $L^{\infty}$.

\subsection{Architecture details} 

We specify the architectures of the 1D networks in \Cref{tab:sm-arch-1d}. We performed polynomial projection with the same number of degrees of freedom as SUPNs.

\begin{table}
    \centering
    \begin{tabular}{c|c|c|c}
        & SUPN & DNN & KAN \\ \hline
        \multirow{2}{*}{$N$ (width)} & 3, 5, 9, 18, & \multirow{2}{*}{2, 3, 5, 9, 12} & \multirow{2}{*}{[6]} \\ 
        & 27, 35, 40 & & \\
        \hline 
        \multirow{2}{*}{$M$ (depth)} & 3, 5, 9, 18,  & \multirow{2}{*}{2, 3, 5, 9, 12} & \multirow{2}{*}{[5]} \\ 
        & 27, 35, 40 & & \\ \hline
    \end{tabular}
    \caption{Network architectures for 1D examples.}
    \label{tab:sm-arch-1d}
\end{table}

\section{Supplementary 2D results}

\subsection{$L^\infty$ best-approximation error}

\begin{figure}
    \centering
    \includegraphics[width=\textwidth]{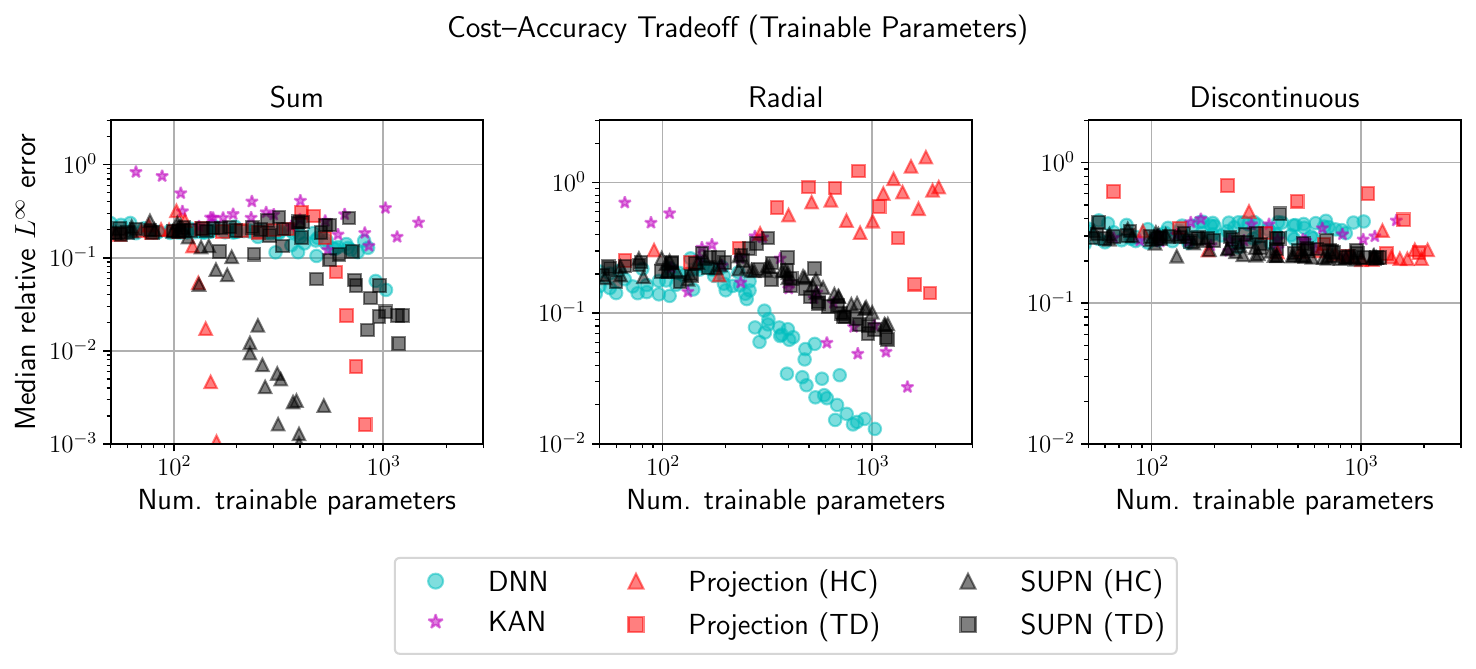}
    \caption{Relative $L^{\infty}$ errors in 2D.}
    \label{fig:sm-best-approx-2d-L-inf}
\end{figure}

$L^\infty$ convergence on the 2D examples (\cref{fig:sm-best-approx-2d-L-inf}) is qualitatively similar to the $L^2$ case, with two exceptions. First, uniform approximation of a discontinuous function by polynomials is error-prone. Second, the $L^\infty$ accuracy of projection on the radial example has seriously degraded from the $L^2$ setting, but it appears that (for $\Lambda_{TD}$, at least) accuracy begins to improve after $P \approx 1000$. This is consistent with the lack of tensor-product structure in the radial target function.

\subsection{$L^2$ robustness}

\begin{figure}
    \centering
    \includegraphics[width=\textwidth]{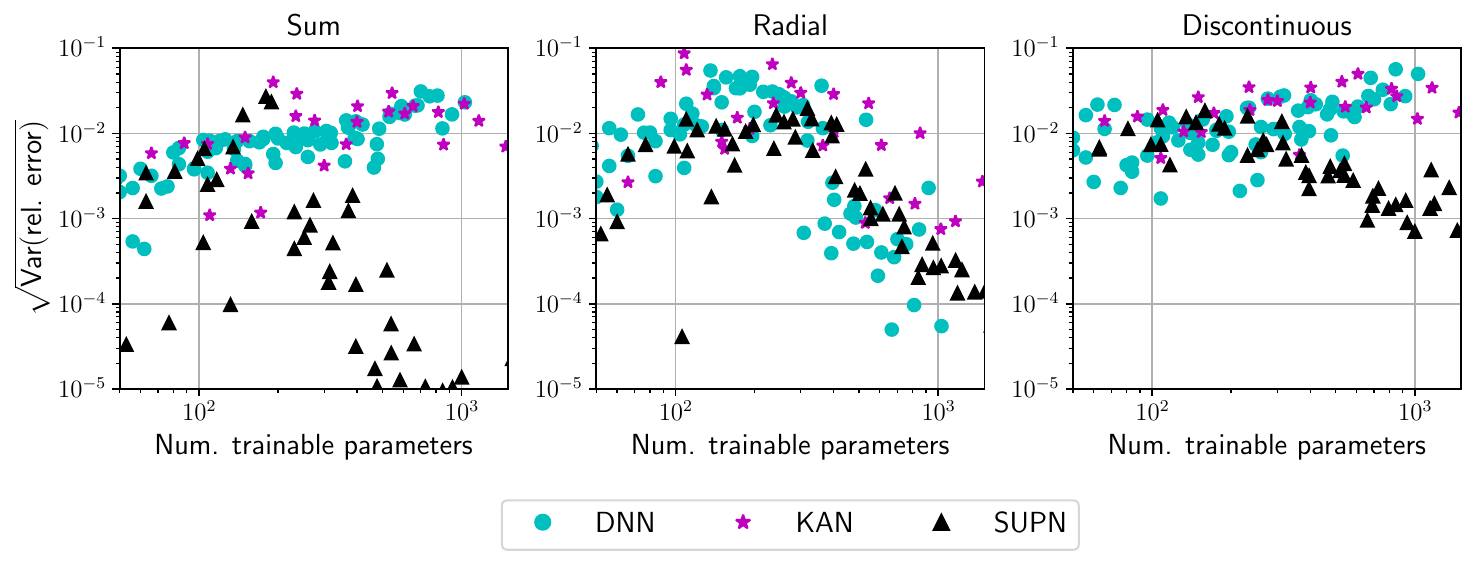}
    \caption{Standard deviation of relative error over 5 realizations of initial network weights for 2D examples. To reduce clutter, only the better-performing of $\Lambda_{\textrm{TD}}$ and $\Lambda_{\textrm{HC}}$ are shown for SUPN.}
    \label{fig:sm-training-robustness-2d}
\end{figure}

\Cref{fig:sm-training-robustness-2d} shows that for $P \gtrsim 100$, SUPNs have between 5-100x less variability in their performance than DNNs and KANs at the same error tolerance. These results comport with those for the one-dimensional examples. Compared to the 1D robustness plots, we do observe an early {\em increase} in variability up to $P \approx 100$. In this regime, networks have sufficiently few parameters to yield reliable predictions that are nonetheless poor.

\subsection{Finite sampling}

\begin{figure}
    \centering
    \includegraphics[width=\textwidth]{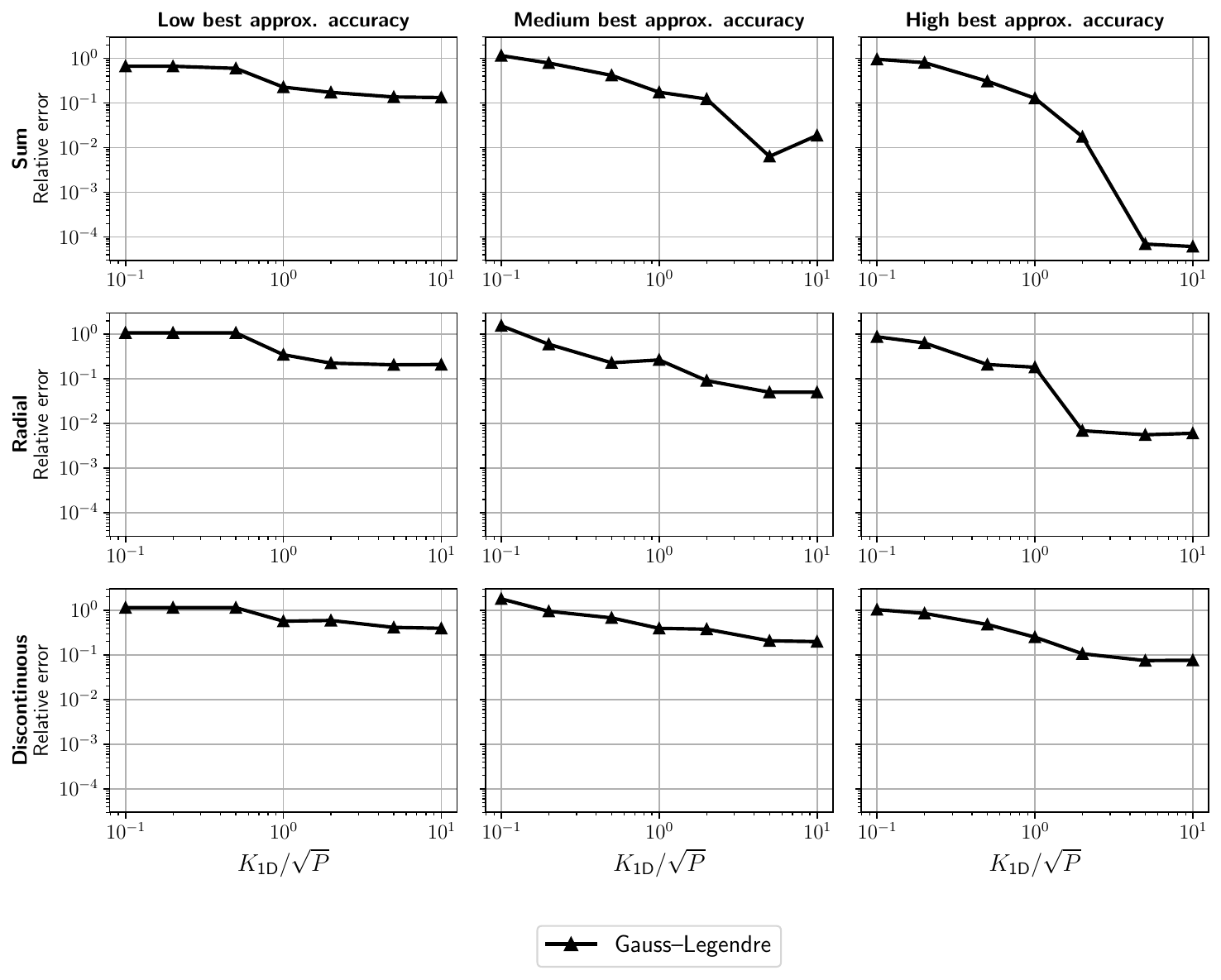}
    \caption{Finite-sampling error on the 2D examples.}
    \label{fig:sm-finite-sampling-supn-2d}
\end{figure}

\Cref{fig:sm-finite-sampling-supn-2d} shows the results of a finite sampling experiment. The horizontal axis is the ratio of the univariate grid size to the square-root of the number of network parameters. The largest ratio needed to obtain best-approximation accuracy is $K_{\text{1D}} \approx 5\sqrt{P}$. Since our networks have $P < 1600$, then our tensor-product Gauss--Legendre rule with 200 nodes in each dimension is sufficient.

\subsection{Architecture details}

\Cref{tab:sm-arch-2d} describes the architecture for the 2D targets. Here, $L$ refers to the contour of either $\Lambda_{\text{HC}}(L)$ or $\Lambda_{\text{TD}}(L)$.

\begin{table}
    \centering
    \begin{tabular}{c|c|c|c|c}
        & SUPN (HC) & SUPN (TD) & DNN & KAN \\ \hline
        \multirow{2}{*}{$N$ (width)} & 1, 3, 5, 7, & 1, 3, 5, 7, & \multirow{2}{*}{[10]} & \multirow{2}{*}{[5]} \\ 
        & 9, 11, 13, 15 & 9, 11, 13, 15 & & \\
        \hline 
        \multirow{2}{*}{$L$ (depth)} & 0,  3,  7, 11,  & 0, 2, 4, 6, & \multirow{2}{*}{[10]} & \multirow{2}{*}{[5]} \\ 
        & 16, 22, 28, 34  & 8, 10, 12, 14 & & \\ \hline
    \end{tabular}
    \caption{Network architectures for 2D examples.}
    \label{tab:sm-arch-2d}
\end{table}

\section{Supplementary 10D results}

\subsection{Architecture details}
\Cref{tab:sm-arch-10d} describes the architecture for the 10D targets.

\begin{table}
    \centering
    \begin{tabular}{c|c|c|c}
        & SUPN (HC) & DNN & KAN \\ \hline
        $N$ (width) & 5, 10, 15, 20 & 20, 40, 60 & 5, 10, 15 \\ \hline 
        $L$ (depth) & [7]  & [4] & [5] \\ \hline
    \end{tabular}
    \caption{Network architectures for 10D example.}
    \label{tab:sm-arch-10d}
\end{table}

\subsection{Noisy observations}

\begin{figure}
    \centering
    \includegraphics[width=0.5\textwidth]{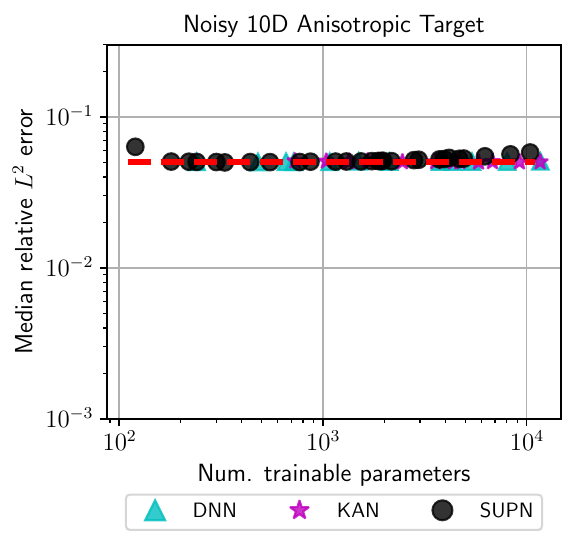}
    \caption{Accuracy from training on noisy data.}
    \label{fig:sm-10d-noisy}
\end{figure}

\Cref{fig:sm-10d-noisy} shows the error arising from training on noise-corrupted samples. In this example, the observations were 
$$
y_i = f(\bm x_i)(1 + \epsilon_i),
$$
where $\epsilon \sim \mathcal{N}(0, 0.05^2)$, so that the relative noise has a standard deviation of 5\%. All methods quickly stagnate at the $5 \times 10^{-2}$ error level (red dashed line).

\section{Best-approximation predictions} \label{sec:sm-best-approx}

In this section, we plot the predictions corresponding to the 1D step function and continuous Rastrigin targets. These plots provide a qualitative complement to the best-approximation error plots in our results. We sweep over $(N, M) \in [10] \times [10]$. For KANs and DNNs, $N$ denotes network width, and $M$ denotes network depth. 

\subsection{Step functions} \label{ssec:sm-stepwise}
\begin{figure}[h!]
    \centering
    \includegraphics[width=.9\textwidth]{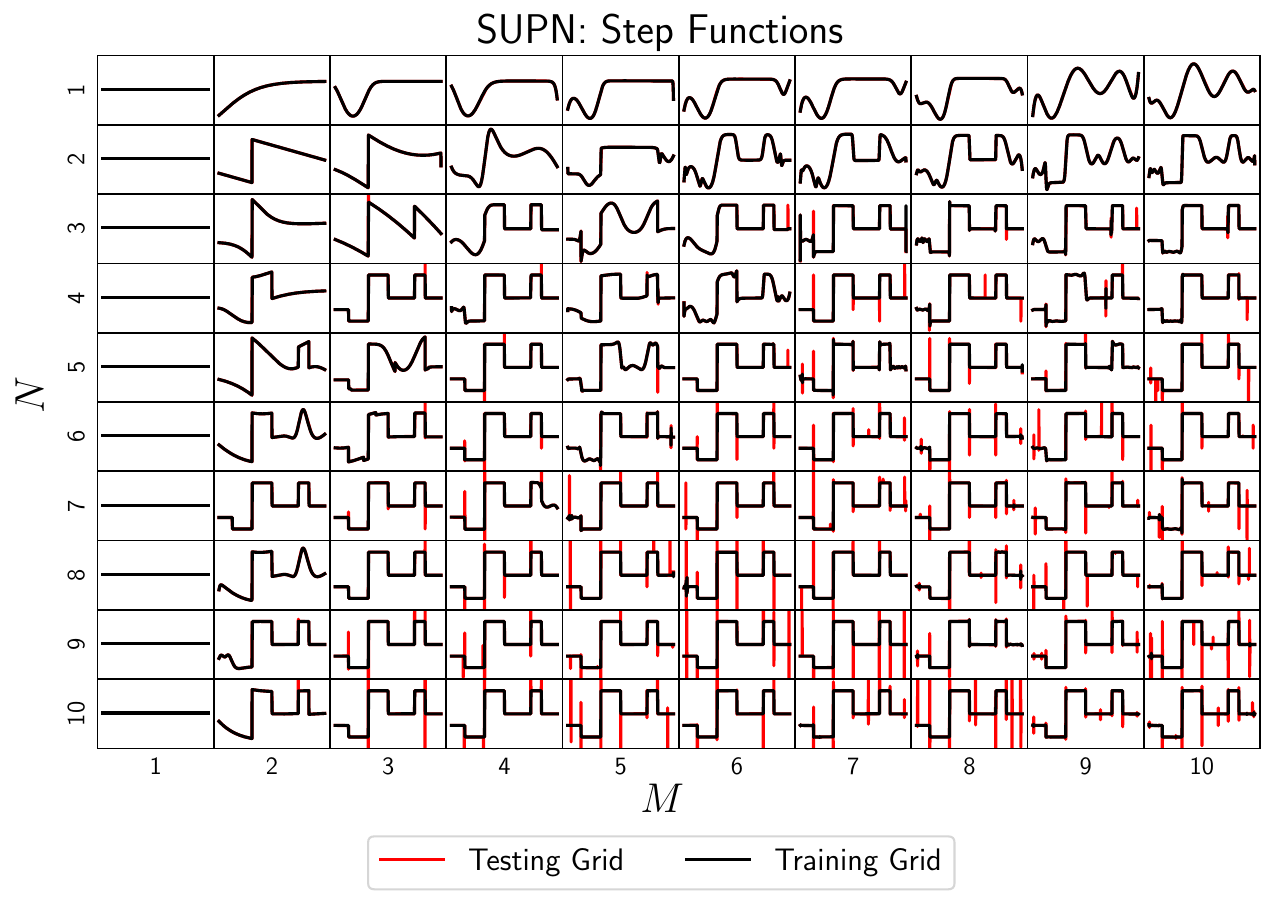}
    \caption{{Plot of best approximation with SUPN on the step functions over $(N,M)$ hyperparameter sweep.}}
    \label{fig:sm-prediction_plots_supn_step}
\end{figure}

\begin{figure}[h!]
    \centering
    \includegraphics[width=.9\textwidth]{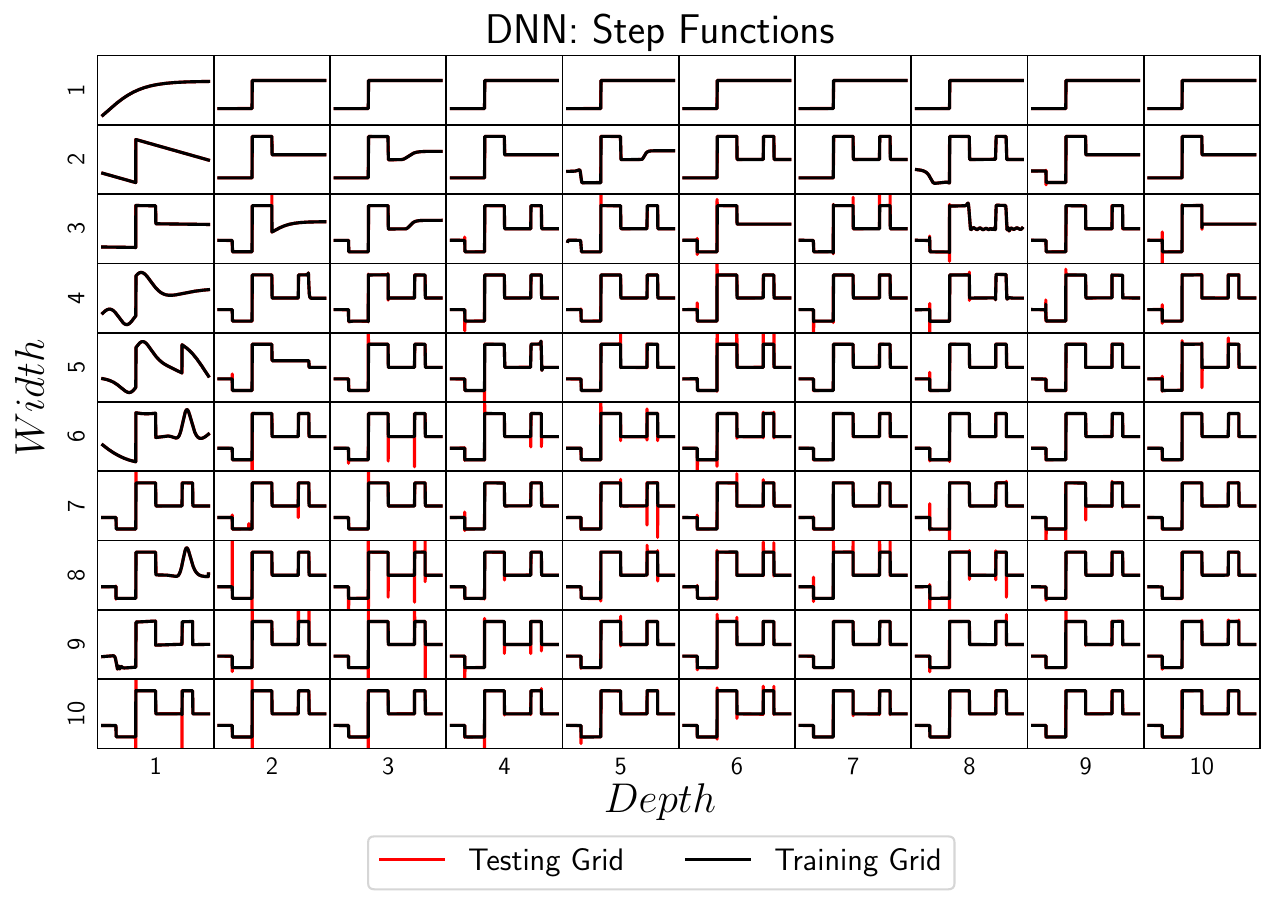}
    \caption{Plot of best approximation with DNN on the step functions over $(Width,Depth)$ hyperparameter sweep.}
    \label{fig:sm-prediction_plots_dnn_step}
\end{figure}

\begin{figure}[h!]
    \centering
    \includegraphics[width=.9\textwidth]{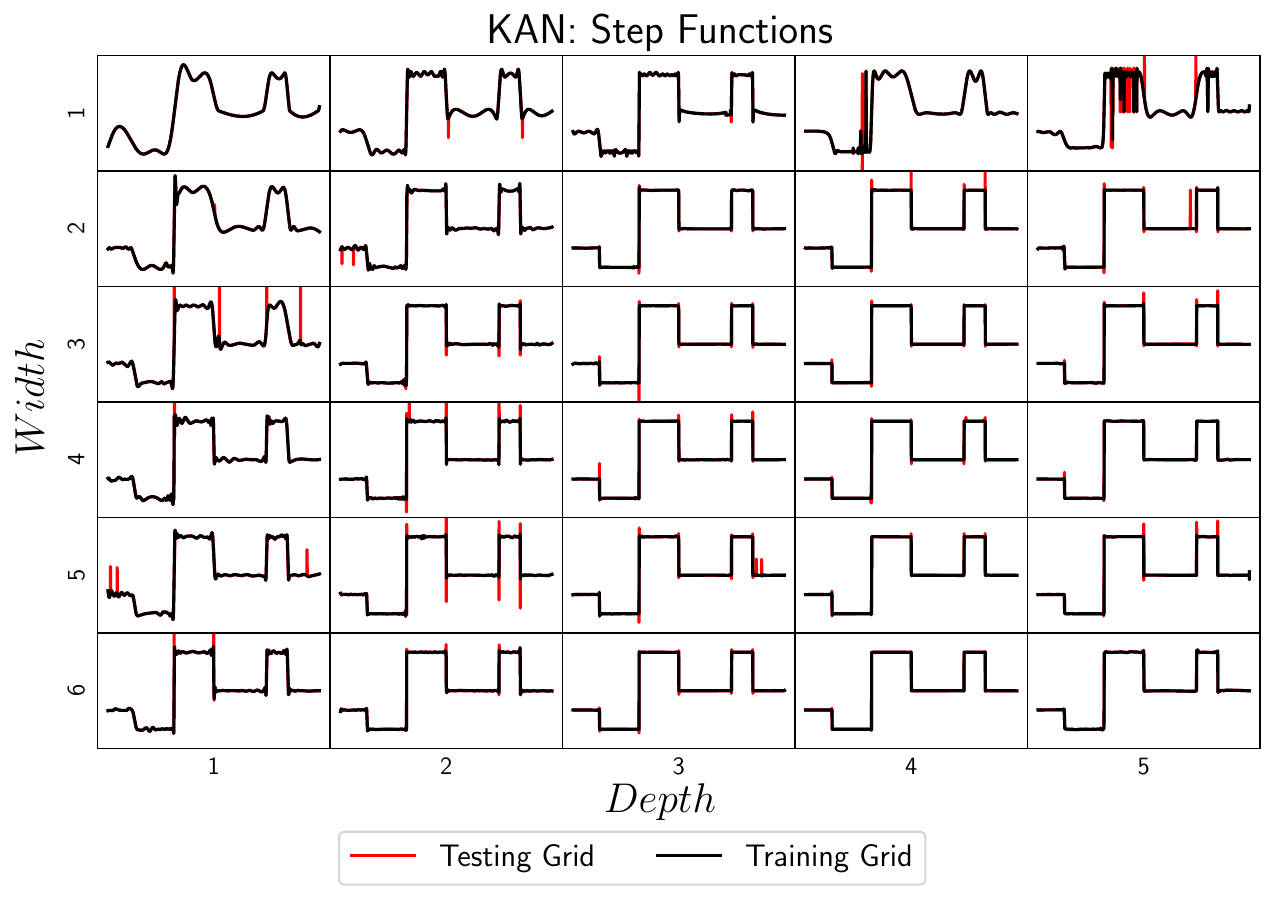}
    \caption{{Plot of best approximation with KAN on the step functions over $(N,M)$ hyperparameter sweep.}}
    \label{fig:sm-prediction_plots_kan_step}
\end{figure}

\begin{figure}[h!]
    \centering
    \includegraphics[width=.9\textwidth]{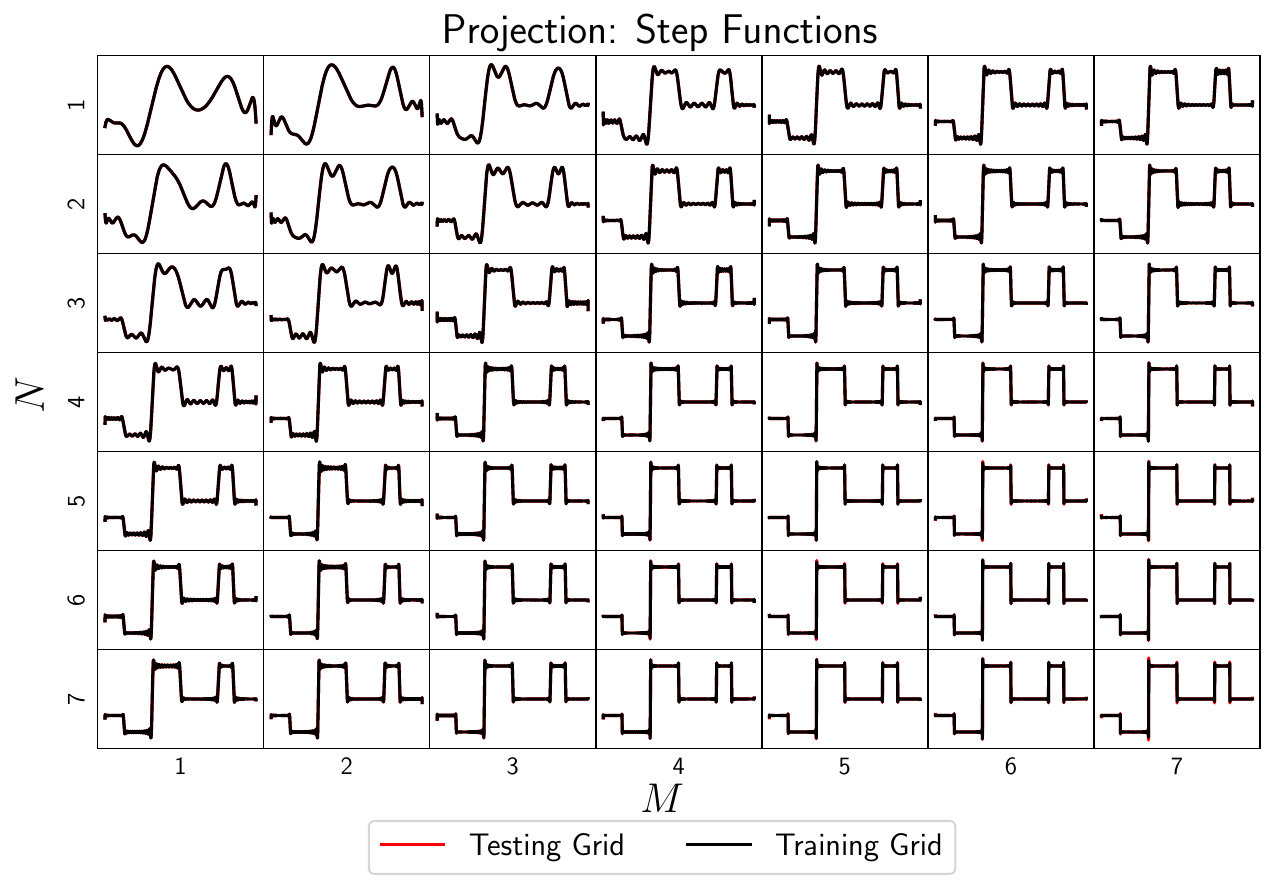}
    \caption{Plot of best approximation with polynomial projection on the step functions over $(N,M)$ hyperparameter sweep.}
    \label{fig:sm-prediction_plots_proj_step}
\end{figure}

\Cref{fig:sm-prediction_plots_supn_step,fig:sm-prediction_plots_dnn_step,fig:sm-prediction_plots_kan_step,fig:sm-prediction_plots_proj_step} demonstrate that the assumption of continuity on the target function is truly necessary for $L^\infty$ convergence. All methods can exhibit persistent spikes over very small sub-intervals of $\Omega$ at the discontinuities, analogous to Gibbs effects. Note that we distinguish between the training and testing grid. These spikes are not present on the training grid, since the error of the approximation is include din the loss and therefore minimized at that grid. However, at finer resolutions during testing, there is no constraint (it is necessary to not have a constraint to fit the discontinuities in the first place during training). To fit the discontinuities at the training grid, the basis functions become extreme which can randomly induce spikes in the underlying approximation which is seen in $L^\infty$ but not $L^2$ since they contribute very little in that norm. This is a limitation of all methods, but is particularly bad for SUPNs, and could be a future research direction to mitigate the effect while still maintaining the ability to fit discontinuities at the training grid level. 

\clearpage
\subsubsection{Zoomed-in Discontinuity Approximations}
\begin{figure}[h!]
    \centering
    \includegraphics[width=.9\textwidth]{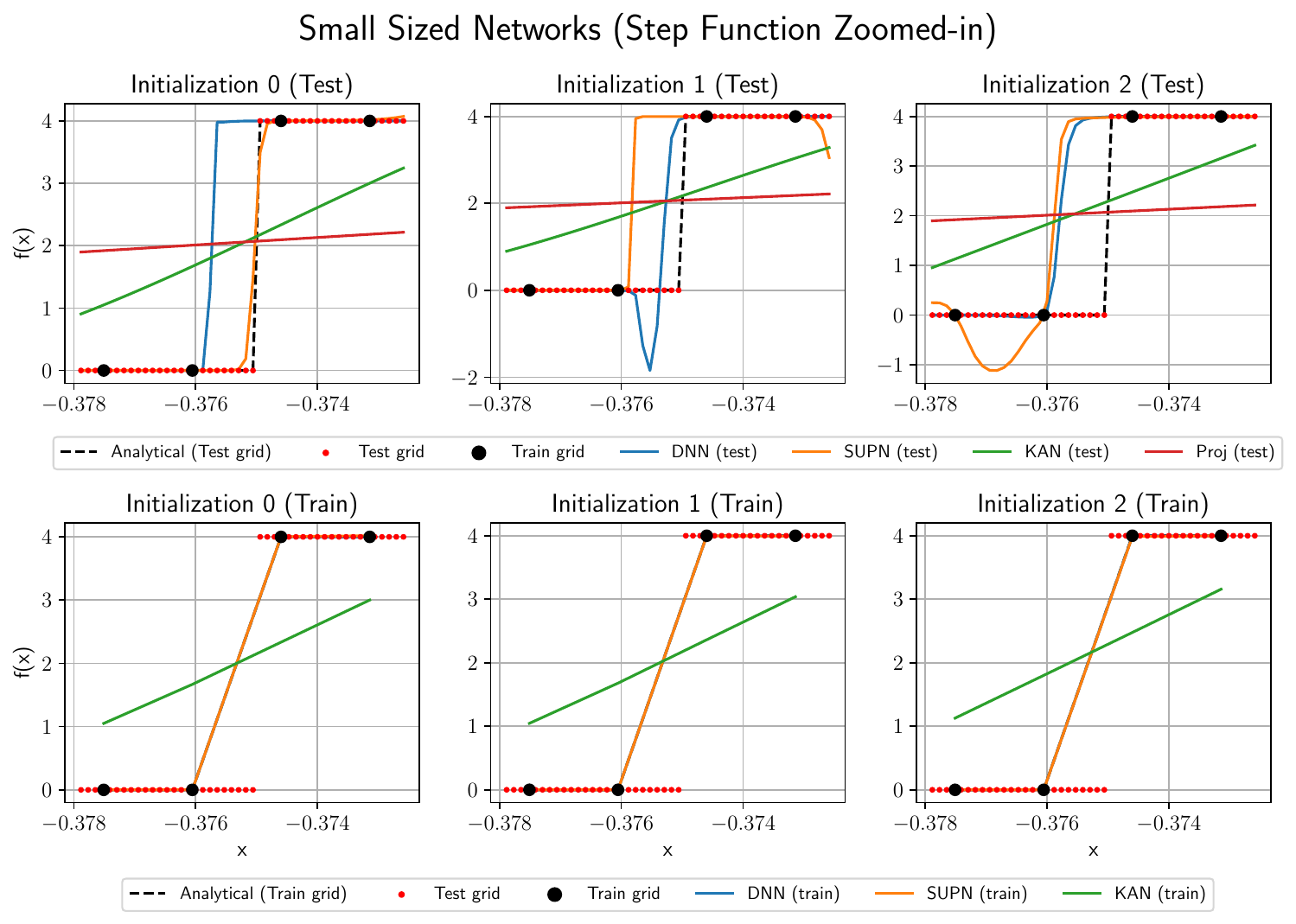}
    \caption{{Plot of zoomed in discontinuity at $x = -0.375$ for the step function target. The network approximations shown are drawn from the \textbf{5th largest} width/depth or M/N models shown in \Cref{fig:sm-prediction_plots_supn_step,fig:sm-prediction_plots_dnn_step,fig:sm-prediction_plots_kan_step}, i.e., along the diagonal, for three different model initializations.}}
    \label{fig:sm-discont_small}
\end{figure}

\begin{figure}[h!]
    \centering
    \includegraphics[width=.9\textwidth]{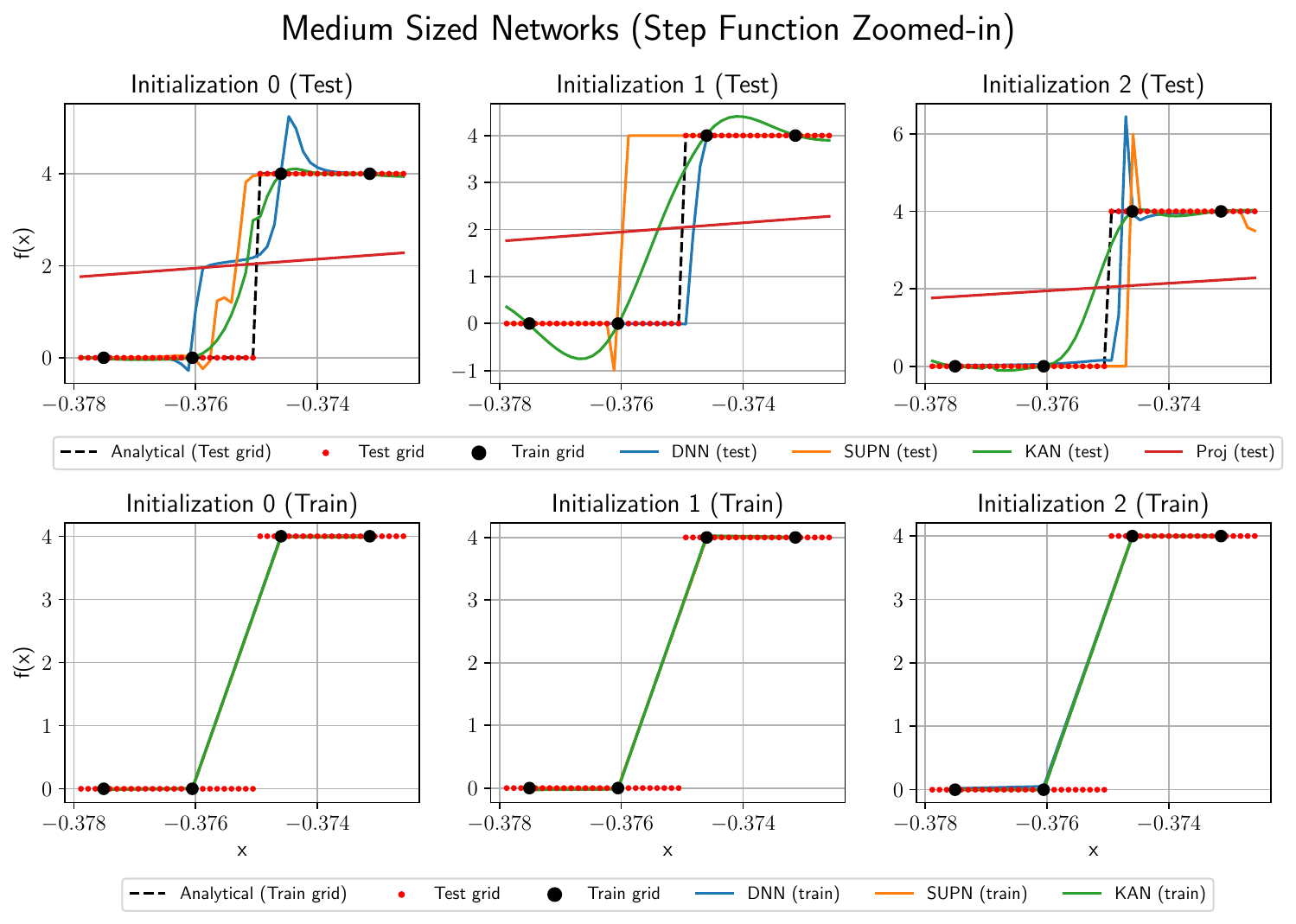}
    \caption{Plot of zoomed in discontinuity at $x = -0.375$ for the step function target. The network approximations shown are drawn from the \textbf{3rd largest} width/depth or M/N models shown in \Cref{fig:sm-prediction_plots_supn_step,fig:sm-prediction_plots_dnn_step,fig:sm-prediction_plots_kan_step}, i.e., along the diagonal, for three different model initializations.}
    \label{fig:sm-discont_medium}
\end{figure}

\begin{figure}[h!]
    \centering
    \includegraphics[width=.9\textwidth]{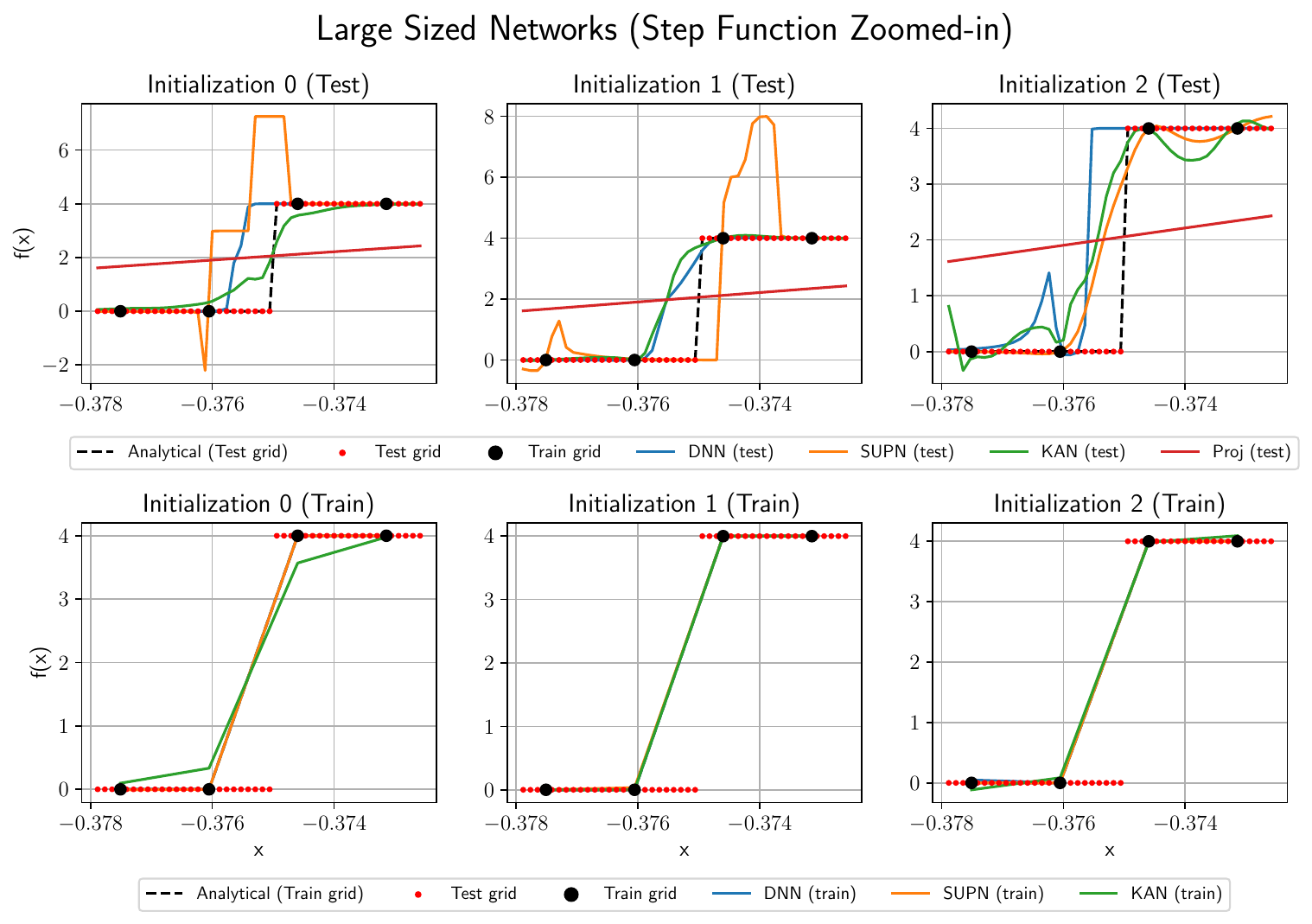}
    \caption{{Plot of zoomed in discontinuity at $x = -0.375$ for the step function target. The network approximations shown are drawn from the \textbf{largest} width/depth or M/N models shown in \Cref{fig:sm-prediction_plots_supn_step,fig:sm-prediction_plots_dnn_step,fig:sm-prediction_plots_kan_step}, i.e., along the diagonal, for three different model initializations.}}
    \label{fig:sm-discont_large}
\end{figure}

\Cref{fig:sm-discont_small,fig:sm-discont_medium,fig:sm-discont_large} provide a more nuanced perspective to the discontinuity approximation and how it drives the lack of $L_{\infty}$ convergence in \Cref{fig:sm-best-approx-1d-L-inf}. Consider the initialization $0$ results in \Cref{fig:sm-discont_small} for the DNN on the test discretization. Despite the approximation perfectly fitting the training points, the method has no knowledge of the analytical location of the discontinuity, so it is misaligned with the target which is observable at higher sampling. Because of this, the $L_{\infty}$ error for the DNN in this case is $4$. Regardless of the spiking behavior observed in all NN-based models (which can be more clearly seen in \Cref{fig:sm-discont_small,fig:sm-discont_medium,fig:sm-discont_large}), they have no chance of ever converging in $L_{\infty}$ due to simply lacking the exact location of the discontinuity at higher discretizations. Other noteworthy observations from these plots is how smooth polynomial projection is compared to the NN-based methods when zoomed in, as well as the fact KAN sometimes fails to pass through the training points as seen in \Cref{fig:sm-discont_large} initialization 0. 

\subsection{Continuous Rastrigin} \label{ssec:sm-rast-cont}
\begin{figure}[h!]
    \centering
    \includegraphics[width=.9\textwidth]{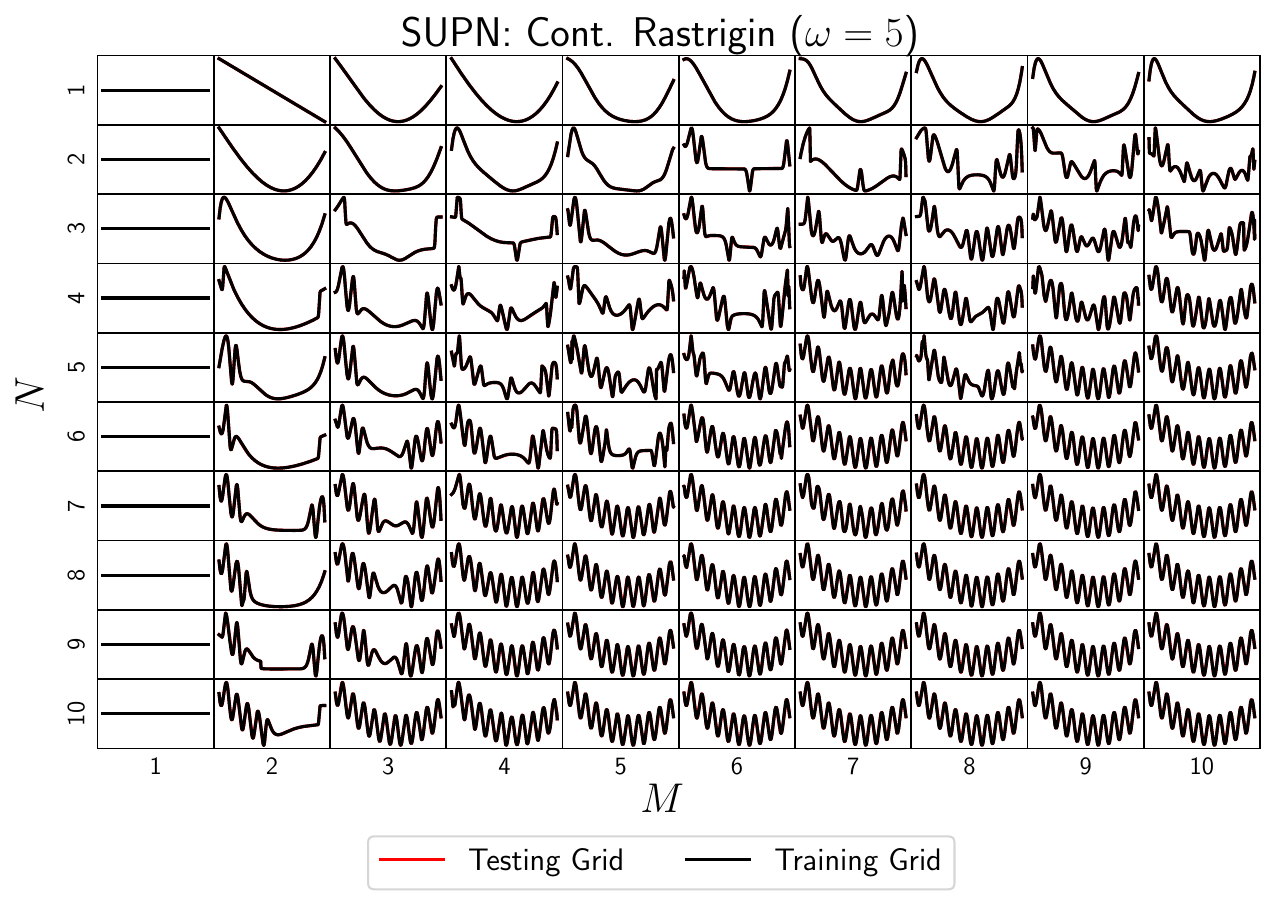}
    \caption{{Plot of best approximation with SUPN on the continuous Rastrigin function over $(N,M)$ hyperparameter sweep.}}
    \label{fig:sm-prediction_plots_supn_rast_cont}
\end{figure}

\begin{figure}[h!]
    \centering
    \includegraphics[width=.9\textwidth]{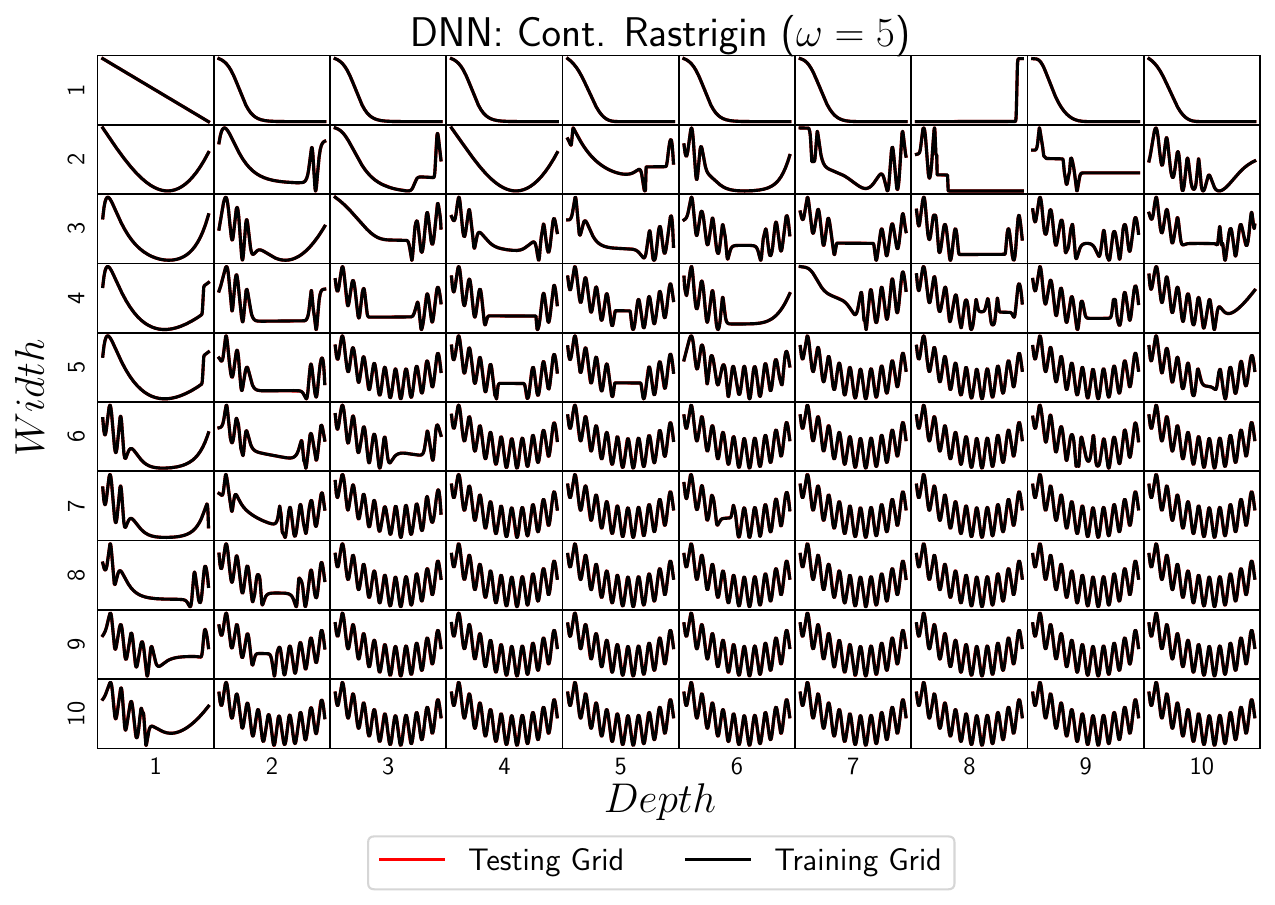}
    \caption{Plot of best approximation with DNN on the continuous Rastirgin function over $(Width,Depth)$ hyperparameter sweep.}
    \label{fig:sm-prediction_plots_dnn_rast_cont}
\end{figure}

\begin{figure}[h!]
    \centering
    \includegraphics[width=.9\textwidth]{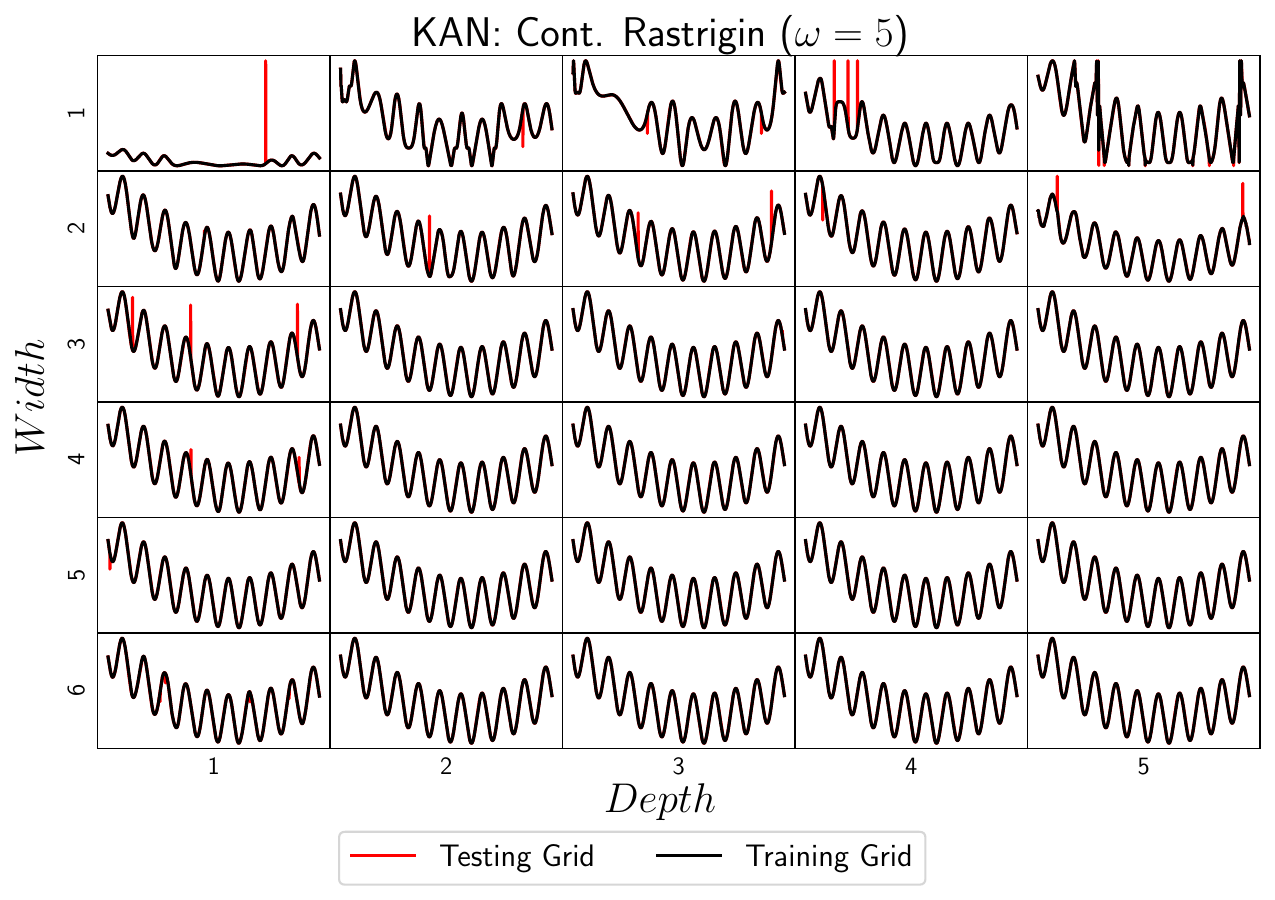}
    \caption{Plot of best approximation with KAN on the continuous Rastirgin function over $(Width,Depth)$ hyperparameter sweep.}
    \label{fig:sm-prediction_plots_kan_rast_cont}
\end{figure}

\begin{figure}[h!]
    \centering
    \includegraphics[width=.9\textwidth]{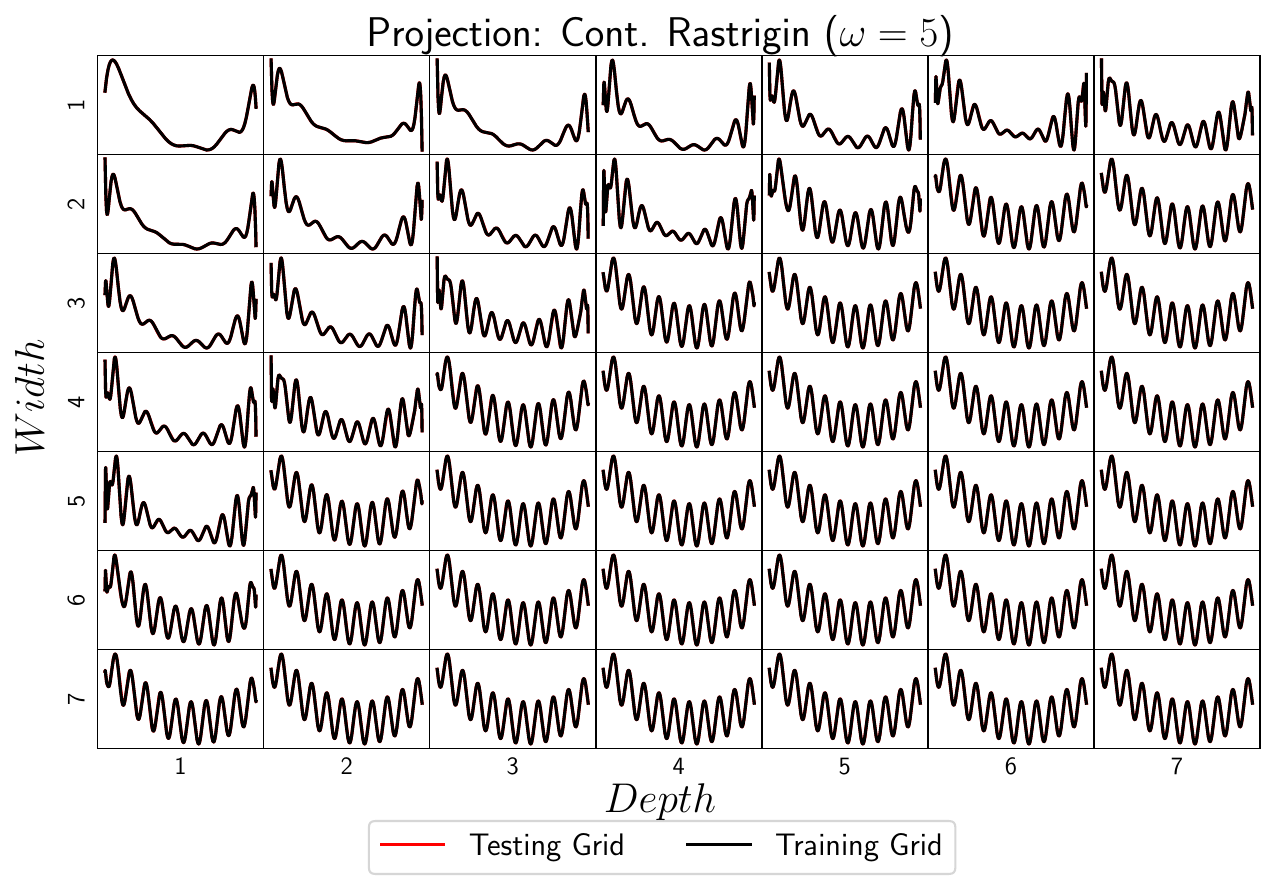}
    \caption{Plot of best approximation with polynomial projection on the continuous Rastrigin function over $(N,M)$ hyperparameter sweep. Each grid cell uses $N(M+1)$ terms, i.e. the same as an SUPN.}
    \label{fig:sm-prediction_plots_proj_rast_cont}
\end{figure}

\Cref{fig:sm-prediction_plots_supn_rast_cont,fig:sm-prediction_plots_dnn_rast_cont,fig:sm-prediction_plots_kan_rast_cont,fig:sm-prediction_plots_proj_rast_cont} visually show the approximation methods as a function of architecture size at both the training and testing discretizations. Unlike \Cref{ssec:sm-stepwise}, we see very little generalization error between the training and testing discretizations, and spiking is not induced at higher discretizations. This gives credence to the idea that the NN-based models learn extreme basis functions to fit the discontinuity more accurately than polynomial projection. For continuous functions, this phenomena is not observed and models converge with respect to their expressability under proper optimization. 

\clearpage

\section{Proof of Sobolev-norm SUPN accuracy bounds} \label{sec:sm-sobolev-theory}
We provide the proof of \cref{prop:poly-approx} for $k > 0$ and $\epsilon_\Lambda(f) > 0$. This proof is conceptually the same as the $k=0$ proof, but involves more opaque constants. By definition of $\epsilon_{\Lambda}(f)$, there is some polynomial $\tilde{q} \in \P_\Lambda$ satisfying \cref{eq:qtilde}.  Let $\tilde{q} = \sum_{\vec{m} \in \Lambda} \alpha_{\vec{m}} T_{\vec{m}}$, and define 
\begin{align*}
  R \coloneqq \|\tilde{q}\|_{W^{k,\infty}(\Omega)} = \max_{|\bm{\beta}| \leq k} \sup_{x \in \Omega} \left|\frac{\partial^{\bm{\beta}}}{\partial \bm{x}^{\bm{\beta}}} \tilde{q}(x)\right| < \infty.
\end{align*}
If $R = 0$, set $c_1 = 0$ in \eqref{eq:ansatz_multidim} with $N = 1$ to achieve $f_{N,\Lambda} = q = 0$, achieving \eqref{eq:supn-q}. Otherwise, consider $R > 0$. 

With $\sigma(y) = \tanh y$, we have that,
\begin{align*}
  \sigma^{(n)}(0) = \left\{\begin{array}{rl} 0, & n \textrm{ even} \\
  \frac{2^{n+1} (2^{n+1} - 1) B_{n+1}}{n+1}, & n \textrm{ odd} \end{array}\right.
\end{align*}
where $B_n$ is the $n$th Bernoulli number.
An additional property of $\sigma$ we require is that there are absolute numbers, $(c_n, d_n)$, such that,
\begin{align*}
  \left| \sigma^{(n)}(y) - \sigma^{(n)}(0) \right| &\leq c_n |y|, & |y| &\leq d_n \leq 1.
\end{align*}
For example, the above is true for any function with a bounded derivative of order $n+2$, which includes the $\tanh$ function. To make computations simpler, we use the looser but sufficient estimate,
\begin{align*}
  \left| \sigma^{(n)}(y) \right| &\leq a_n, & |y| \leq \tilde{d}_n,
\end{align*}
where $a_n = \max\{1, \sigma^{(n)}(0)\}$, where $\tilde{d}_n \in (0, d_n)$.

To compute an order-$k$ derivative of an SUPN, Fa\'{a} di Bruno's formula states that, for a function $q$ with sufficient differentiable smoothness,
\begin{align*}
  \frac{d^k}{d x^k} \sigma(q(x)) &= \sum_{\ell =0}^k \sigma^{(\ell)}(q(x)) B_{k,\ell}\left(q^{(1)}, \ldots, q^{(k-\ell+1)}\right) \\
  &= \sigma'(q(x)) B_{k,1} \left(q^{(1)}, \ldots, q^{(k)}\right) + \sum_{\substack{\ell =0\\\ell\neq 1}}^{k} \sigma^{(\ell)}(q(x)) B_{k,\ell}\left(q^{(1)}, \ldots, q^{(k-\ell+1)}\right),
\end{align*}
where $B_{k,\ell}$ is the $k$-variate (exponential) incomplete Bell polynomial of degree $\ell$. We note that $B_{k,1}(s_1, \ldots, s_k) = s_k$ for all $k$, and $B_{k,0} = 0$ for any $k > 0$, so that our formula simplifies to,
\begin{align*}
  \frac{d^k}{d x^k} \sigma(q(x)) &= \sigma'(q(x)) q^{(k)}(x) + \sum_{\ell =2}^k \sigma^{(\ell)}(q(x)) B_{k,\ell}\left(q^{(1)}, \ldots, q^{(k-\ell+1)}\right),
\end{align*}
We will need suprema of the incomplete Bell polynomials over a compact isotropic hypercube of edgelength $R$:
\begin{align*}
  \overline{B}_k &\coloneqq \max_{\ell =0, \ldots, k} \|B_{k,\ell}(z_1, \ldots, z_k)\|_{L^\infty([-R,R]^k)}.
\end{align*}
A final property of the Bell polynomials we need is that for $r \geq 2$, $B_{r,\ell}(z_1, \ldots, z_r)$ is homogeneous of order $\ell$, i.e., $B_{r,\ell}(c z_1, \ldots, c z_r) = c^\ell B_{r,\ell}(z_1, \ldots, z_r)$.

For simplicity, we take the multi-index $\bm{\beta}$ specifying the derivative to be a (multiple of a) cardinal unit vector, i.e., $\bm{\beta} = r \bm{e}_s$, where $r \in [1, k]$, and $\bm{e}_s \in \R^d$ is the cardinal unit vector in direction $s$. Then $\partial_{\bm{\beta}} f = \partial_s^r f$. For any $r \in [1, k]$,
\begin{align*}
  \frac{\partial^r}{\partial x_s^r} \sigma(q(\bm{x})) - q^{(r,s)}(\bm{x}) &= \underbrace{\left(\sigma'(q(x)) - 1\right) q^{(r)}(x)}_{\mathrm{(A)}} \\
  &\hspace{2em} + \underbrace{\sum_{\ell=2}^{r} \sigma^{(\ell)}(q(x)) B_{r,\ell}\left(q^{(1,s)}, \ldots, q^{(r-\ell+1,s)}\right)}_{\mathrm{(B)}}
\end{align*}
where for shorthand we've written $q^{(r,s)}$ to denote the $r$-th partial derivative of $q$ in the $s$th variable. Now, define $q$ as,
\begin{align*}
  q(x) = \frac{1}{S} \tilde{q}(x), \qquad S &= \max\{ (\delta \epsilon_{\Lambda}(f))^{-1/k}, (\delta \epsilon_{\Lambda}(f))^2\} \ \times \\
  &  \qquad \max\left\{1, \frac{R}{\min\{\tilde{d}_1, \ldots, \tilde{d}_k\}}, R^{3/2}, k \left( \overline{B}_k \, \max\{a_2, \ldots, a_k \} \right) ^{1/k}\right\}
\end{align*}
Then:
\begin{align*}
  \left|\mathrm{(A)}\right| &\leq \left| \sigma'(q(x)) - 1 \right| R/S \leq |q(x)|^2 R/S \leq R^3/S^3 \leq \delta \epsilon_\Lambda(f) \frac{1}{S}\\
  \left|\mathrm{(B)}\right| &\leq \sum_{\ell=2}^r |\sigma^{(\ell)}(q(x))| \left|B_{r,\ell}\left(q^{(1,s)}, \ldots, q^{(r-\ell+1,s)}\right)\right| \\
  &\leq \sum_{\ell=2}^r a_{\ell} \left|B_{r,\ell}\left(q^{(1,s)}, \ldots, q^{(r-\ell+1,s)}\right)\right| \\
  &\leq \sum_{\ell=2}^r a_{\ell} S^{-\ell} \left|B_{r,\ell}\left(\tilde{q}^{(1,s)}, \ldots, \tilde{q}^{(r-\ell+1,s)}\right)\right| \\
  &\leq \sum_{\ell=2}^r a_{\ell} S^{-\ell} \overline{B}_r \\
  &\leq \frac{1}{S} \sum_{\ell=2}^k \frac{a_{\ell} \overline{B}_k}{S^{\ell-1}} \, .
\end{align*}
If $\delta \epsilon_{\Lambda}(f) \geq 1$, then 
\begin{align*}
\left|\mathrm{(A)}\right| \leq 1/S \, \sum_{\ell=2}^k \frac{a_{\ell} \overline{B}_k}{(k \,\overline{B}_k \max\{a_2, \dots, a_k\})^{(\ell-1)/k}} \, (\delta \epsilon_{\Lambda}(f))^{(\ell-1)/k} \leq (1/S) 
\end{align*}
The same estimate holds for all $\bm{\beta}$ satisfying $|\bm{\beta}| \leq k$ by taking iterated derivatives in each variable.
Now consider the $(N,M) = (1,|\Lambda|)$ SUPN in \eqref{eq:ansatz_multidim}, with the parameter assignment
\begin{align*}
  a_{1,\vec{m}} &= \frac{\alpha_{\vec{m}}}{S}, & c_1 &= S,
\end{align*}
so that $f_{N,\Lambda}(\bm{x}) = S \sigma(q(\bm{x}))$. We have shown, for any $\vec{x} \in \Omega$ and $| \vec{\beta}| \leq k$,
\begin{align*}
  \left| \frac{\partial^{\bm{\beta}}}{\partial \bm{x}^{\bm{\beta}}} f_{N, \Lambda}(\vec{x}) - \frac{\partial^{\bm{\beta}}}{\partial \bm{x}^{\bm{\beta}}} \tilde{q}(\bm{x}) \right| = S \left| \frac{\partial^{\bm{\beta}}}{\partial \bm{x}^{\bm{\beta}}} \sigma(q(\bm{x})) - \frac{\partial^{\bm{\beta}}}{\partial \bm{x}^{\bm{\beta}}} q(\bm{x}) \right| \leq \delta \epsilon_{\Lambda}(f) \, ,
\end{align*}
which completes the proof.

\end{document}